%% file: main.tex
\input{kth_thesis_style_papers}

\usepackage[headheight=15pt]{geometry}  
\usepackage[intlimits]{mathtools}       
\usepackage{graphicx}                   
\usepackage{booktabs}                   
\usepackage{multirow}                   
\usepackage[normalem]{ulem}             
\useunder{\uline}{\ul}{}                
\usepackage{appendix}                   
\usepackage{xspace}
\usepackage{amsthm}                     
\usepackage{hyperref}                   
\usepackage{natbib}                     
\usepackage{lscape}                     
\hypersetup{
    pdftitle={Target Population Synthesis using CT-GAN},
    pdfauthor={Tanay Rastogi},
    colorlinks=true,
    linkcolor=black,
    citecolor=black,
    filecolor=black,
    urlcolor=black
}

\begin{document}

\title{Target Population Synthesis using CT-GAN}
\author{Tanay Rastogi and Daniel Jonsson}
\date{}
\maketitle
\pagenumbering{arabic}

\begin{abstract}
\noindent Agent-based models used in scenario planning for transportation and urban planning usually require detailed population information from the base as well as target scenarios. These populations are usually provided by synthesizing fake agents through deterministic population synthesis methods. However, these deterministic population synthesis methods face several challenges, such as handling high-dimensional data, scalability, and zero-cell issues, particularly when generating populations for target scenarios. This research looks into how a deep generative model called Conditional Tabular Generative Adversarial Network (CT-GAN) can be used to create target populations either directly from a collection of marginal constraints or through a hybrid method that combines CT-GAN with Fitness-based Synthesis Combinatorial Optimization (FBS-CO). The research evaluates the proposed population synthesis models against travel survey and zonal-level aggregated population data. Results indicate that the stand-alone CT-GAN model performs the best when compared with FBS-CO and the hybrid model. CT-GAN by itself can create realistic-looking groups that match single-variable distributions, but it struggles to maintain relationships between multiple variables. However, the hybrid model demonstrates improved performance compared to FBS-CO by leveraging CT-GAN’s ability to generate a descriptive base population, which is then refined using FBS-CO to align with target-year marginals. This study demonstrates that CT-GAN represents an effective methodology for target populations and highlights how deep generative models can be successfully integrated with conventional synthesis techniques to enhance their performance.

\end{abstract}

\newpage
\section{Introduction}\label{Introduction}

Scenario planning is a strategic method used by urban planners to create various plausible future scenarios based on current trends and uncertainties, helping decision-makers anticipate challenges and craft adaptable strategies. Agent-based models (ABM) enhance this process by simulating interactions among individual agents, capturing their behaviors and decision-making processes. These methods are particularly valuable as they allow for the exploration of complex systems and dynamics through detailed modeling of individual actions and choices.

These ABMs typically require comprehensive data regarding the population of an area, including individuals' social and household characteristics. In scenario planning, these traits are needed for both the base and target populations. However, such detailed information about the whole population is usually unavailable, owing to issues such as privacy concerns and technical and financial constraints in data gathering. Instead, statistical authorities in many countries have made available micro-samples of individual-level data from the whole population. In addition to these micro-samples, aggregated marginal information on a regional or zonal level is also available from the Bureau of Statistics. Using these various partial views of the population, researchers can synthesize a more comprehensive representation of the actual population using population synthesis algorithms. 

The goal of population synthesis is to find the best way to use different data sources to create agents in social and geographical spaces that are very close to the underlying population structure and meet certain scenario criteria set by the user, such as the correlation structure and control totals \citep{Guo2007PopulationBehavior, Axhausen2010PopulationArt, Ma2016AnPopulations}.

The research conducted by \cite{Rich2018Large-scaleDenmark, Borysov2019HowSynthesis} highlights that the population synthesis methods typically involve three steps: 1) the initial step involves using the base population to represent various combinations of attributes, 2) the fitting stage involves estimating the weighting factors for the base population using the control tables to construct the representative target population, and 3) the allocation stage involves the generation and assignment of synthetic agents to ABM transport models. By following the steps above, the target population can be made directly from the available micro-sample. However, most researchers will generate a larger pool of synthetic agents for the base population in the pre-processing step to use all the aggregate information that is available. Then this resulting synthetic pool is used for the allocation stage to generate the target population. 

In the population synthesis literature, there has been a trend toward statistical learning (SL) for the generation of synthetic populations, in place of deterministic methods like Iterative Proportional Fitting and Combinatorial Optimization. Multiple research from \cite{Farooq2013SimulationSynthesis, Sun2015ASynthesis, Borysov2019HowSynthesis, Garrido2020PredictionModelling, Kim2023ASynthesis} used various types of SL models, including one based on Gibbs sampling called Markov Chain Monte Carlo (MCMC), bayesian networks, and deep generative models like variational autoencoder and generative adversarial networks. However, a major drawback of these methods is that they fail to satisfy the conditional distribution required for the target population. For this reason, most of these previously mentioned studies have only focused on generating a base population without considering how such samples can be aligned with target scenarios.

This paper analyzes the use of SL-based population synthesis model for generating target population, specifically employing the Conditional Tabular Generative Adversarial Network (CT-GAN), proposed by \cite{Xu2019ModelingGAN}, to create synthetic populations based on user-defined aggregated marginals for target scenarios. The study contributes by connecting target-population synthesis with deep generative modeling, demonstrating how CT-GAN can be adapted for this problem. 

We also analyze the advantages and disadvantages of combining CT-GAN with conventional methods such as the Fitness-based Synthesis Combinatorial Optimization (FBS-CO) model. In this hybrid approach, the CT-GAN is used to create a diverse base population that can be synthesized into a target population using FBS-CO. The hybrid model takes advantage of CT-GAN's skill in handling complex data to overcome problems that traditional models face, like heterogeneity in data, scalability issues, zero-cell issues, and small sample sizes, which could result in more accurate representations of target populations.

The remainder of this article is structured in the following manner: Section \ref{Literature} provide a brief background on various population synthesis algorithms and the contribution of this research, Section \ref{Method} provides a detailed description of the CT-GAN model, FBS-based CO model and how we use these models for generating synthetic population in this study. To evaluate and access the performance of the population synthesis methods, we utilize data from a travel survey and aggregated zonal data. The experimental setup, metric evaluation, results, and discussions on these are provided in Section \ref{Experiments}. Finally, in Section \ref{Conclusion}, the article concludes by summarizing the analysis.

\section{Literature Review}\label{Literature}
According to \cite{Sun2018ASynthesis, FabriceYameogo2020ComparingPopulation}, numerous methods have been proposed for population synthesis, categorized into three main approaches: Synthetic Reconstruction (SR), Combinatorial Optimization (CO), and Statistical Learning (SL). The SR approach, including methods like Iterative Proportional Fitting (IPF) and Iterative Proportional Update, integrates sample data and aggregate statistics to calculate weights that indicate each sample agent's representativeness in a specific zone. CO methods, such as Genetic Algorithms and Fitness-based Synthesis, also utilize sample and aggregate data to select household combinations that best match the marginals. In contrast, SL methods like Markov Chain Monte Carlo (MCMC), Bayesian methods, and Deep Generative Models (DGM) focus solely on the sample data, estimating probabilities for each attribute combination based on the joint distribution.

Introduced by \cite{Deming1940OnKnown}, IPF is one of the most important SR method used for population synthesis and has been successfully implemented in numerous studies. Research from \cite{Pritchard2012AdvancesSimultaneously, E.Ramadan2020AModels} provides a comprehensive overview of various studies that use IPF for population synthesis. Several researchers have employed IPF for generating both baseline and target populations. \cite{Rich2018Large-scaleDenmark} proposed a large-scale population synthesis framework for Denmark involving target harmonization, matrix fitting, post-simulation of households and agents, and re-weighting of the final population, using this framework to generate a population for target-year 2015 based on the 2010 population as initial seed. 

Similarly, \cite{Predhumeau2023ACanada} developed a hybrid approach called Quasi-random Integer Sampling of IPF (QISI), which combines IPF and QIS by constructing a distribution with IPF and then sampling the integral population without replacement. In their study, they first synthesized the population for 2016 using a micro-sample and subsequently used it as seed for generating populations for futures year using the aggregates from 2021, 2023, and 2030. Other studies have examined the accuracy and scalability of target-year population synthesis. \cite{Ma2016AnPopulations} presents an empirical assessment of the accuracy of target-year populations synthesized with different seed-data and control tables with varying noise using IPF and CO methods. \cite{Saadi2018InvestigatingApproach} contributes to the state-of-the-art by comparing the effect of scalability on the quality of synthetic populations generated by IPF and MCMC, discussing the interactions between changes in sampling rate and scalability.

The CO method, iteratively replaces households with a new set of individuals and households until a goodness-of-fit indicator converges toward specified stopping criteria. One key advantage of CO methods is that their data requirements remain less restrictive than those for SR methods. Additionally, CO directly generates a list of households that match multiple multilevel controls without needing to create a joint multi-way distribution, which is a limitation of IPF \citep{Ma2015SyntheticValidations}. Studies such as \cite{Williamson1998TheRecords} have used CO-based techniques like synthetic reconstruction and re-weighting to generate synthetic populations. \cite{Ma2015SyntheticValidations} presents the FBS approach for generating synthetic populations that can efficiently handle multilevel controls, demonstrating its feasibility and improved performance compared to traditional IPF methods.

In recent years, SL methods (also called simulation-based algorithms) have gained momentum for generating synthetic populations. These methods focus on the joint distribution of all attributes in the sample by directly estimating a probability for each combination \citep{Sun2018ASynthesis}. Even when specific agents do not exist in the original data, it may still be possible to sample these agents by combining agents from the original dataset. These methods excel in addressing high-dimensional problems, offering better scaling properties and fulfilling the need for more detailed populations. 

As highlighted by research from \cite{Farooq2013SimulationSynthesis, Sun2015ASynthesis}, SL methods effectively address both the lack of heterogeneity raised by SR and CO methods and challenges associated with small sample sizes. Furthermore, SL methods can generate synthetic populations from sample data alone when marginal information is unavailable. However, a major drawback of these methods is their failure to satisfy the conditional distribution required for target-year population. For this reason, most previous studies \cite{Farooq2013SimulationSynthesis, Sun2015ASynthesis, Sun2018ASynthesis, Borysov2019HowSynthesis, Garrido2020PredictionModelling, Kim2023ASynthesis} have focused only on generating base-year populations without considering how such samples can be aligned with future targets. 

As highlighted by \cite{FabriceYameogo2020ComparingPopulation}, combining conventional SR or CO -methods with SL-methods might be the most effective approach for generating synthetic populations. If micro-sample data are neither comprehensive (lacking observations for each type of individual in the actual population) nor representative of the population, yet marginals are available, an SL method can first be applied to construct a suitable base population. Subsequently, SR or CO methods can be used to create a target population that matches the marginals.

The research mentioned above have primarily focused on using SR and CO for generating target populations. To the author's knowledge, there has been no study focused on using SL methods to synthesize target populations with user-defined aggregated marginal constraints. Additionally, no study has explored the hybrid methods that combine SR/CO methods with SL methods and analyzing the advantages and disadvantages of using this combination to generate a target population. Hence, the main contributions of this study are summarized below.
\begin{itemize}
    \item Proposes an SL-based deep generative model, called the CT-GAN, that can effectively synthesize a target population based on given conditions while ensuring a degree of accuracy at least equivalent to the methods described in existing literature.
    \item Analyzes the advantages and disadvantages of a hybrid method (CO + SL) for synthesizing a target population, providing valuable insights for future research in this field.
\end{itemize}

\section{Methodology}\label{Method}
This study proposes a deep generative model-based method called Conditional Tabular GAN (CT-GAN), which is trained on a small micro-sample to generate synthetic population data. The trained CT-GAN model is applied in two distinct ways:
\begin{enumerate}
\item \textit{Stand-alone} - CT-GAN directly synthesizes a target population using conditional marginals provided for the target area.
\item \textit{Hybrid} - CT-GAN generates a descriptive base population that subsequently serves as input for creating a target population via the FBS-CO method.
\end{enumerate}

\begin{figure}[htbp]
\centering
{{\includegraphics[scale=0.85]{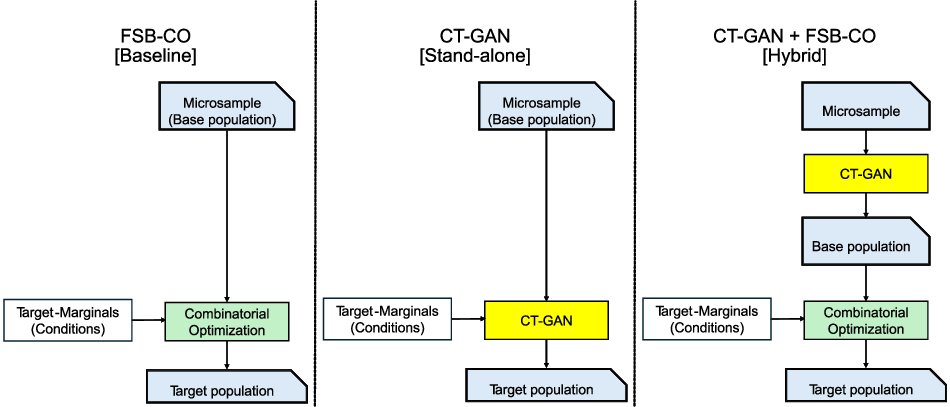}}}
\caption{Conceptual framework illustrating the three approaches for target population synthesis employed in this study.}
\label{fig:flowchart}
\end{figure}

The CT-GAN model in this study is trained on a micro-sample without any location-specific information. Subsequently, we utilize this model to synthesize both the base and target populations. For comparative evaluation, we also generate a population using the FBS-CO method alone, which serves as our baseline for analyzing and comparing both the stand-alone CT-GAN and the hybrid approach. Figure \ref{fig:flowchart} illustrates the conceptual framework for synthesizing target populations using these three different approaches: baseline, stand-alone, and hybrid. All methods are evaluated based on their ability to accurately replicate the statistical distribution of the actual population. Detailed descriptions of the CT-GAN methodology and the FBS-CO method are provided in Sections \ref{sec:CTGAN} and \ref{sec:FBS-CO}, respectively

\subsection{Conditional Tabular GAN}\label{sec:CTGAN}
The CT-GAN model, proposed by \cite{Xu2019ModelingGAN}, is a DGM-based method for modeling tabular data distribution and sampling rows from it. CT-GAN offers several advantages over other DGMs in modeling tabular data. As highlighted by Xu et al. in their study, it addresses common issues in population synthesis by introducing mode-specific normalization and implementing architectural changes. Additionally, it tackles data imbalances through a conditional generator and employs training-by-sampling techniques. These enhancements enable CT-GAN to effectively model both discrete and continuous columns simultaneously, manage multi-modal non-Gaussian values within continuous columns, and mitigate severe imbalances in categorical columns.

CT-GAN generates synthetic populations by training a data synthesizer network, called Generator G, which learns from a table of data T. In our case, T represents data from a population of agents $n = 1, 2, 3..., N$ with $K$ discrete columns of agent attributes $\{D_1, ..., D_K\}$, i.e., variables representing agents' individual and household characteristics. Each column is considered a random variable. These agent attributes follow an unknown joint distribution $P(D_K)$. The goal of the CT-GAN model is to estimate this joint distribution of attributes to generate a table of synthetic data $T_{syn}$ that approximates the true joint distributions of attributes across a real population.

A major advantage of using CT-GAN is that it addresses the issue of highly imbalanced categorical columns, which leads to mode collapse in vanilla GANs. CT-GAN uses conditional generators to solve this problem. It samples a discrete column $D_i*$ and one of its values $k*$ and then generates a sample $\hat{r}$ based on that condition, i.e., $\hat{r} \sim P_G(row|D_i* = k*)$. This is achieved through conditional vector and training-by-sample methods, each discussed in detail in the paper by \cite{Xu2019ModelingGAN}. Once trained, this unique feature of CT-GAN helps synthesize tabular data with specific conditions, which is useful for generating target populations.

For training the model, $G$ is initiated with a draw from a standard normal latent variable, $Z$, and the conditional vector as inputs. The output of the generator is the synthesized rows, $\hat{r}$. The second network, the discriminator network $D$, receives real and synthetic samples, with corresponding conditional vectors, as input and scores them. The objective of $D$ is to determine whether the information it receives comes from real or synthetic data. The training process continues until the $D$ network can no longer distinguish between generated and real data.

Once trained, the CT-GAN model employs a \textit{reject sampling approach} to generate synthetic data that adheres to specific conditions based on aggregated marginals for selected attributes. The process begins with the generator network producing synthetic data rows using random noise vectors and conditional vectors as inputs. These generated samples are then evaluated against predefined conditions or constraints. Samples that meet the specified conditions are "accepted" and included in the synthetic dataset, while those that fail to meet the conditions are "rejected" and discarded. This iterative process continues until a sufficient number of samples meeting the desired conditions are obtained, ensuring that the resulting synthetic data closely aligns with the target population characteristics while maintaining the overall distribution learned by the CT-GAN model.

\begin{figure}[htbp]
\centering
{{\includegraphics[scale=0.5]{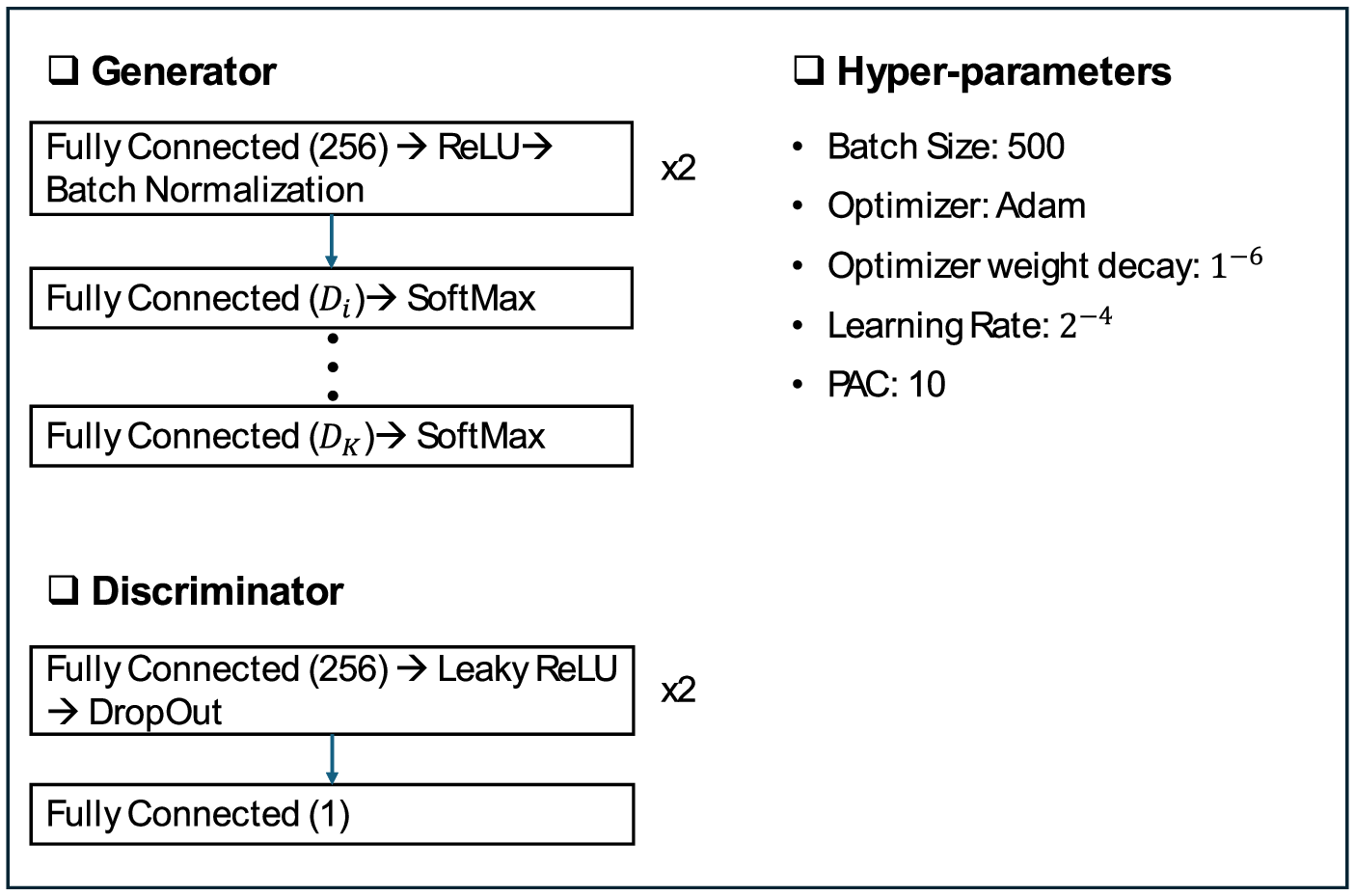}}}
\caption{CT-GAN network architecture and hyper-parameters used for training the model, as described in \cite{Xu2019ModelingGAN}.}
\label{fig:ct-gan}
\end{figure}

In our research, we adopt the original network architecture for both the generator and discriminator, along with their corresponding hyper-parameters, as outlined in the original CT-GAN paper. The model is trained using the Wasserstein GAN loss function with gradient penalty. The specific model architecture and hyper-parameters are illustrated in Figure \ref{fig:ct-gan}.

\subsection{Combinatorial Optimization}\label{sec:FBS-CO}
The Combinatorial Optimization (CO) is one of the popular population synthesis methods that is used for synthesizing target population. The Fitness‐Based Synthesis Combinatorial Optimization (FBS-CO) method is a way to “synthesize” a target population by selecting (or “drawing”) individuals from a micro-sample so that the aggregated characteristics (such as age, gender, or other demographic attributes) closely match the known target distribution. In mathematical terms, the approach can be formulated as a discrete optimization problem with a fitness (or error) function that quantifies the mismatch between the synthetic and target distributions.

In this framework, $x_i$ denotes the number of times candidate $i$ is chosen (allowing for sampling with replacement) from a set of candidate individuals $i = 1, 2, ... n$. Then, the vector $\textbf{x} = (x_1, x_2, ..., x_n)$ must satisfy the desired total population constraint: 
\begin{equation}
     \sum_{i=1}^{n} x_i = N
\end{equation}

An attribute matrix $\textbf{A}$ is defined such that each entry $A_{ki}$ indicates whether candidate $i$ contributes to the $k_{th}$ attribute, and the aggregate distribution of attributes in the synthetic population is given by $\textbf{Ax}$. The method then seeks to minimize a fitness function, called, relative sum of squared Z scores (RSSZ) \citep{Ryan2009PopulationFirms}. The RSSZ is given as:

\begin{equation}
    RSSZ = \sum_{k} SSZ_k
\end{equation}

\begin{equation}
\begin{aligned}
    \text{where} \quad SSZ_k = \sum_{i} F_{ki} (Ax_{ki} - E_{ki})^2 \\
    F_{ki} = 
    \begin{cases} 
        \left( C_k Ax_{ki} \left( 1 - \frac{Ax_{ki}}{N_k} \right)\right)^{-1}, & \text{if } Ax_{ki} \neq 0 \\ 
        \frac{1}{C_k}, & \text{if } Ax_{ki} = 0 
    \end{cases}
\end{aligned}
\end{equation}

$Ax_{ki}$ is the observed (from the subset) count for the $i_{th}$ cell of the $k_{th}$ tabulation (characteristic); $E_{ki}$ is the expected (known) count for the $i_{th}$ cell of the $k_{th}$ tabulation; $N_k$ is the total count of tabulation $k$; and $C_k$ is the 5\% $\chi^2$ critical value for tabulation k (with $n-1$ degrees of freedom, for a table with n cells). Through an iterative combinatorial optimization process—typically involving small adjustments such as swapping counts between candidates to reduce discrepancies between the synthetic and target data. The RSSZ metric not only measures the magnitude of these discrepancies but also accounts for their statistical significance, ensuring that the synthetic population aligns with the target distributions in a statistically meaningful way.

\section{Case Study}\label{Experiments}
For the purpose of evaluating the proposed method for synthesizing a target population, we employed two distinct types of datasets: travel surveys conducted over different years and zone-level data containing aggregated marginals for each zone. The primary difference between these datasets lies in the level of detail and location-specific information. Travel surveys provide agent-level information but lack any location-specific identifiers. This agent-level data is subsequently aggregated to produce marginals, which serve as input to the models under evaluation. In contrast, the zonal-level dataset comprises only aggregated marginals to specific attributes without any individual-level details. The specific details of these datasets are described in Sections \ref{sec:data_travel_survey} and \ref{sec:data_zonal_level}.

In this study, we train the proposed CT-GAN model using travel survey data. The CT-GAN is designed to learn from the individual-specific attributes present in the travel survey data without any location-specific information. The details of the model training procedure are outlined in Section \ref{sec:model_eval}. Following this, Section \ref{sec:travel_survey} presents the evaluation results, focusing on the CT-GAN’s ability to synthesize future travel surveys based on learned patterns from the historical data.

Subsequently, Section \ref{sec:ct-gan-zonal} provides a comparative analysis of three population synthesis models, described in Section \ref{Method}, using the aggregated zonal-level data. The performance of each model is assessed in terms of their ability to generate disaggregated population data when supplied only with aggregated marginals for specified zones.

\subsection{Dataset}
\subsubsection{Travel Survey dataset} \label{sec:data_travel_survey}
The travel survey for this study comes from the national travel behavior survey conducted Swedish Transport Analysis. The dataset is from two different sources. The first one is called Riks-RVU 2005–2006, and that serves are the micro-sample used for training the CT-GAN as well as the base year population for FBS-CO and hybrid models. The survey was conducted over a year, from October 2005 to September 2006 (\cite{RES0506}). The second travel survey data is from national travel survey covering the six consecutive years from 2011 to 2016 (\cite{Holmstrom2025TravelSurvey}). Both these survey encompasses individuals aged 6-84, drawn from a stratified sample of the Swedish total population register. Each row in the survey entry represents a weighted sample, post-stratified based on year, region, age, and sex. We focused solely on attributes that characterize a prototypical citizen from the travel survey, as detailed in the Table \ref{tab:attributeTable}, deliberately excluding any location-specific data such as home or work zone information.

\begin{table}[htbp]
\centering
\caption{Individual attributes available in the travel survey data.}
\label{tab:attributeTable}
\resizebox{0.8\columnwidth}{!}{%
\begin{tabular}{@{}llll@{}}
\toprule
\multicolumn{1}{c}{\textbf{Attribute}} & \multicolumn{1}{c}{\textbf{Type}} & \multicolumn{1}{c}{\textbf{Description}} & \multicolumn{1}{c}{\textbf{Info}} \\ \midrule
AGE & Categorical & Age & Categories: 7 \\
SEX & Categorical & Gender & Categories: 2 \\
DRVLIC & Boolean & Driving license possession & Bool: True/False \\
LIFECATG & Categorical & Type of living status & Categories: 7 \\
EDULEVEL & Categorical & Level of education & Categories: 7 \\
WORK & Categorical & Type of work situation & Categories: 3 \\ \bottomrule
\end{tabular}%
}
\end{table}

The aggregate target-marginals that are used as input to models for synthesizing future population are created by aggregating these survey data over age and gender attributes. Table \ref{tab:travel-aggr} presents these marginals for each target year, along with the total number of agents per year. For reference, the marginals for the base year 2005/06 are also included.

\begin{table}[htbp]
\centering
\caption{Aggregated marginals on Age and Gender from Travel Survey data.}
\label{tab:travel-aggr}
\resizebox{0.8\columnwidth}{!}{%
\begin{tabular}{@{}cccccccll@{}}
\multicolumn{1}{l}{} & \multicolumn{2}{c}{\textbf{2005/06}} & \multicolumn{2}{c}{\textbf{2011}} & \multicolumn{2}{c}{\textbf{2012}} & \multicolumn{2}{c}{\textbf{2013}} \\ \midrule
\textbf{Age / Sex} & \textbf{f} & \textbf{m} & \textbf{f} & \textbf{m} & \textbf{f} & \textbf{m} & \multicolumn{1}{c}{\textbf{f}} & \multicolumn{1}{c}{\textbf{m}} \\ \midrule
0-6 & 143 & 142 & 95 & 75 & 37 & 37 & \multicolumn{1}{c}{25} & \multicolumn{1}{c}{23} \\
7-15 & 1943 & 1798 & 635 & 590 & 463 & 395 & \multicolumn{1}{c}{238} & \multicolumn{1}{c}{218} \\
16-19 & 566 & 525 & 566 & 533 & 184 & 179 & \multicolumn{1}{c}{100} & \multicolumn{1}{c}{80} \\
20-24 & 846 & 733 & 561 & 427 & 210 & 171 & \multicolumn{1}{c}{107} & \multicolumn{1}{c}{88} \\
25-44 & 3782 & 3785 & 2085 & 2332 & 801 & 786 & \multicolumn{1}{c}{429} & \multicolumn{1}{c}{458} \\
45-64 & 3541 & 3635 & 2500 & 2805 & 979 & 1071 & \multicolumn{1}{c}{558} & \multicolumn{1}{c}{632} \\
65+ & 1363 & 1498 & 1796 & 1971 & 817 & 855 & \multicolumn{1}{c}{533} & \multicolumn{1}{c}{580} \\ \midrule
\textbf{Total Pop.} & \multicolumn{2}{c}{24300} & \multicolumn{2}{c}{16971} & \multicolumn{2}{c}{6985} & \multicolumn{2}{c}{4069} \\ \midrule
\multicolumn{1}{l}{} & \multicolumn{1}{l}{} & \multicolumn{1}{l}{} & \multicolumn{1}{l}{} & \multicolumn{1}{l}{} & \multicolumn{1}{l}{} & \multicolumn{1}{l}{} &  &  \\
\multicolumn{1}{l}{} & \multicolumn{2}{c}{\textbf{2014}} & \multicolumn{2}{c}{\textbf{2015}} & \multicolumn{2}{c}{\textbf{2016}} &  &  \\ \cmidrule(r){1-7}
\textbf{Age / Sex} & \textbf{f} & \textbf{m} & \textbf{f} & \textbf{m} & \textbf{f} & \textbf{m} &  &  \\ \cmidrule(r){1-7}
0-6 & 50 & 41 & 27 & 29 & 37 & 32 &  &  \\
7-15 & 394 & 362 & 297 & 268 & 307 & 288 &  &  \\
16-19 & 293 & 236 & 114 & 124 & 117 & 98 &  &  \\
20-24 & 322 & 261 & 146 & 130 & 117 & 93 &  &  \\
25-44 & 1227 & 1282 & 468 & 486 & 416 & 386 &  &  \\
45-64 & 1600 & 1825 & 653 & 690 & 525 & 610 &  &  \\
65+ & 1633 & 1729 & 752 & 798 & 646 & 694 &  &  \\ \cmidrule(r){1-7}
\textbf{Total Pop.} & \multicolumn{2}{c}{11255} & \multicolumn{2}{c}{4982} & \multicolumn{2}{c}{4366} &  &  \\ \cmidrule(r){1-7}
\end{tabular}%
}
\end{table}

\newpage
\subsubsection{Aggregated Zonal-level dataset}\label{sec:data_zonal_level}
For this analysis, we used the Population (Befolkning) dataset, which provides aggregated marginals in a three-way frequency table for AGE, SEX, and WORK across Sweden at the SAMS (Small Area Market Statistics)\footnote{Sweden is divided into 9000 statistical areas, known as SAMS, or Small Area Market Statistics.} zonal level for 2005. In this study, we focused exclusively on zones within Umeå's municipal region, resulting in a total of 89 zone blocks. The population in each Umeå-SAMS zone varies considerably, ranging from as few as 3 to as many as 4,424 inhabitants. For example, Table \ref{tab:sams-exp} presents marginals from two zones with the lowest and highest populations, respectively. This significant variation in population size poses an additional challenge for population synthesis models, as they must generate synthetic populations that satisfy the diverse conditional requirements of each zone.

Within this zonal-level dataset, the WORK attribute is a binary variable with two categories: working and not-working. In the training data derived from the travel survey, the WORK attribute initially includes three categories: working, not-working, and part-time. To ensure consistency with the available ground truth, we convert the “part-time” category to “working,” thereby creating a binary classification. This conversion allows the synthetic populations generated by each model to be directly comparable to the observed data and thus suitable for subsequent analysis.

\begin{table}[htbp]
\centering
\caption{Examples of population data for two zones in Umeå-SAMS for year 2005.}
\label{tab:sams-exp}
\resizebox{0.8\columnwidth}{!}{%
\begin{tabular}{@{}ccccc@{}}
\cmidrule(l){2-5}
\multicolumn{1}{l}{} & \multicolumn{4}{c}{\textbf{24800061}} \\ \midrule
\textbf{\begin{tabular}[c]{@{}c@{}}AGE /\\ {[}SEX, WORK{]}\end{tabular}} & \textbf{\begin{tabular}[c]{@{}c@{}}f,\\ not\_working\end{tabular}} & \textbf{\begin{tabular}[c]{@{}c@{}}f, \\ working\end{tabular}} & \textbf{\begin{tabular}[c]{@{}c@{}}m, \\ not\_working\end{tabular}} & \textbf{\begin{tabular}[c]{@{}c@{}}m,\\ working\end{tabular}} \\ \midrule
0-6 & 57 & 0 & 61 & 0 \\
7-15 & 50 & 0 & 63 & 0 \\
16-19 & 47 & 14 & 41 & 6 \\
20-24 & 777 & 267 & 670 & 199 \\
25-44 & 447 & 255 & 729 & 409 \\
45-64 & 35 & 92 & 54 & 75 \\
65+ & 34 & 0 & 37 & 5 \\ \midrule
\multicolumn{1}{l}{\textbf{Total Pop.}} & 1447 & 628 & 1655 & 694 \\ \midrule
\multicolumn{1}{l}{} & \multicolumn{1}{l}{} & \multicolumn{1}{l}{} & \multicolumn{1}{l}{} & \multicolumn{1}{l}{} \\ \cmidrule(l){2-5} 
\multicolumn{1}{l}{} & \multicolumn{4}{c}{\textbf{24800018}} \\ \midrule
\textbf{\begin{tabular}[c]{@{}c@{}}AGE /\\ {[}SEX, WORK{]}\end{tabular}} & \textbf{\begin{tabular}[c]{@{}c@{}}f,\\ not\_working\end{tabular}} & \textbf{\begin{tabular}[c]{@{}c@{}}f,\\ working\end{tabular}} & \textbf{\begin{tabular}[c]{@{}c@{}}m,\\ not\_working\end{tabular}} & \textbf{\begin{tabular}[c]{@{}c@{}}m,\\ working\end{tabular}} \\ \midrule
0-6 & 0 & 0 & 0 & 0 \\
7-15 & 0 & 0 & 0 & 0 \\
16-19 & 0 & 0 & 0 & 0 \\
20-24 & 0 & 0 & 0 & 0 \\
25-44 & 0 & 0 & 0 & 0 \\
45-64 & 0 & 0 & 3 & 0 \\
65+ & 0 & 0 & 0 & 0 \\ \midrule
\multicolumn{1}{l}{\textbf{Total Pop.}} & 0 & 0 & 3 & 0 \\ \bottomrule
\end{tabular}%
}
\end{table}

\newpage
\subsection{Model Evaluation and Discussion} \label{sec:model_eval}
In our analysis, the proposed CT-GAN models is trained on the travel survey data for year 2005/2006. We trained the model using the hyper-parameters specified in Figure \ref{fig:ct-gan} that includes parameters for discriminator and generator, dimension of latent space vector, learning rate and optimizer related parameters. The model underwent training on a GPU cluster comprising two NVIDIA GeForce RTX 3080 units, each with 10GB of memory, for a total of 400 epochs. The loss calculated at each epoch for both discriminator and generator is presented in the plot in APPENDIX \ref{apx:ct_gan_training}. Ultimately, this trained model is used to synthesize the population all analysis.

\subsubsection{Metrics}
In order to evaluate the quality of the generated synthetic population, we perform two levels of checks - single column and multi-dimensional level. The column level check ensures that each attribute individually follows the statistical distribution of the actual population. Subsequently, the evaluation of multi-dimensional joint attribute distributions verifies whether the interactions among attributes present in the actual population are accurately replicated in the synthetic population across all dimensions. Specifically, our evaluation focused on the following two aspects:
\begin{itemize}
    \item How closely the generated data matched the provided conditional marginals?
    \item How well the model captured the distributions of other attributes not explicitly conditioned on?
\end{itemize}

The attribute level check is conducted using two metrics - Total Variation Complement (TVC) and Category Adherence (CA) provided by the SDMetrics library by \cite{Datacebo2024SDMetrics}. 
The TVC metric computes the similarity between an actual attribute and a synthetic attribute in terms of column shapes, i.e., the marginal distribution or 1D histogram of the column. This test calculates the Total Variation Distance (TVD) between the actual and synthetic attributes. To accomplish this, it first computes the frequency of each category value and expresses it as a probability. The TVD statistic compares the differences in probabilities as:
\begin{align}
\label{eq:tvd}
\delta(R,S)=\frac{1}{2}\sum_{\omega\in\eta}^{}\left| R_\omega-S_\omega \right|
\end{align} 
Here, $\omega$ describes all the possible categories in an attribute set $\eta$. Meanwhile, R and S refer to the ground-truth and synthetic frequencies for those categories, respectively. The TVC returns 1-TVD so that a higher score indicates higher quality and is given by:
\begin{align}
\label{eq:tvs_score}
score_{tvc} = 1 - \delta(R,S)
\end{align} 

The CA metric measures whether a synthetic attribute adheres to the same category values as the ground-truth data, meaning the synthetic population should not introduce new category values that are not originally present in the ground-truth population. This metric extracts the set of unique categories present in the ground-truth attribute, denoted as $C_r$. It then identifies the data points in the synthetic data, $s$, that are found in the set $C_r$. The score is calculated as the proportion of these data points compared to all synthetic data points and is given by:
\begin{align}
\label{eq:ca_score}
score_{ca} = \frac{\left | s,s\in C_r \right |}{\left | s \right |}
\end{align}

Similar to other research like \cite{Kim2023ASynthesis, Garrido2020PredictionModelling, Borysov2019HowSynthesis, Saadi2016ForecastingApproaches}, the joint distribution of attributes evaluation is performed by employing the Standardized Root Mean Square Error (SRMSE) as a metric for assessing multi-dimensional distributions. The SRMSE is given by:

\begin{align}
\label{eq:SRMSE}
SRMSE(\pi, \hat{\pi} | k)=\frac{RMSE}{\overline{\pi}}=\frac{\sqrt{\sum_{(i, j)}^{}(\pi_{(i, j)}-\hat{\pi}{(i, j)})^2/N{b}}}{\sum_{(i, j)}\pi_{(i, j)}/N_b}
\end{align}

where $\pi$ and $\hat{\pi}$ are k-joint categorical distributions of ground-truth and synthetic populations, respectively. $N_b$ represents the total number of possible category combinations, and $k$ is the number of attributes in the joint distribution.

\subsubsection{Evaluation on Travel Survey} \label{sec:travel_survey}
To evaluate the CT-GAN's capability for synthesizing target populations, we applied it to Swedish travel survey data collected over multiple years. To generate the target population for each year, we provided the CT-GAN model with aggregated marginals for AGE and SEX as input conditioning variables, presented in Table \ref{tab:travel-aggr}. As noted earlier, we evaluated the CT-GAN model’s ability to precisely meet predefined AGE and SEX conditions while preserving the complex, realistic inter-dependencies among all population variables.

Table \ref{tab:attribute-level} presents the results of the column-level check for all target synthetic data, evaluated against the corresponding actual target data and is supported by the bar plots for attribute distribution for each year in APPENDIX \ref{apx:travel_survey_plots}. The results demonstrate that the CT-GAN model consistently generates data that exactly match the provided marginals for AGE and SEX across all years, confirming its ability to replicate the actual data for the given aggregated marginals. For attributes that were not explicitly conditioned (DRVLIC, LIFECATG, and WORK), the CT-GAN produces data that closely approximates the actual data, with TVC and CA scores approaching 0.9. However, the model struggles with EDULEVEL, possibly due to insufficient target data for this attribute and the large number of categories involved. Also, the model’s performance declines as the amount of missing data increases, with the worst results observed in 2015 and 2016, where nearly 94\% of the EDULEVEL data are missing. 

\begin{table}[htbp]
\centering
\caption{Attribute level analysis for all CT-GAN synthetic population against corresponding target population.}
\label{tab:attribute-level}
\resizebox{0.9\columnwidth}{!}{%
\begin{tabular}{@{}cccccccccc@{}}
\cmidrule(l){2-10}
 & \multicolumn{3}{c}{\textbf{2011}} & \multicolumn{3}{c}{\textbf{2012}} & \multicolumn{3}{c}{\textbf{2013}} \\ \midrule
\textbf{Attribute} & \textbf{\%Na} & \textbf{TVC} & \textbf{CA} & \textbf{\%Na} & \textbf{TVC} & \textbf{CA} & \textbf{\%Na} & \textbf{TVC} & \textbf{CA} \\ \midrule
Age & 0 & 1 & 1 & 0 & 1 & 1 & 0 & 1 & 1 \\
Sex & 0 & 1 & 1 & 0 & 1 & 1 & 0 & 1 & 1 \\
Work & 0.004 & 0.908 & 1 & 0 & 0.937 & 1 & 0 & 0.935 & 1 \\
DrvLic & 0 & 0.929 & 1 & 0 & 0.922 & 1 & 0 & 0.935 & 1 \\
EduLevel & 0.376 & 0.817 & 1 & 0.369 & 0.801 & 1 & 0.343 & 0.805 & 1 \\
LifeCatg & 0.619 & 0.936 & 1 & 0.631 & 0.951 & 1 & 0.657 & 0.936 & 1 \\ \midrule
\multicolumn{1}{l}{} & \multicolumn{1}{l}{} & \multicolumn{1}{l}{} & \multicolumn{1}{l}{} & \multicolumn{1}{l}{} & \multicolumn{1}{l}{} & \multicolumn{1}{l}{} & \multicolumn{1}{l}{} & \multicolumn{1}{l}{} & \multicolumn{1}{l}{} \\ \cmidrule(l){2-10} 
 & \multicolumn{3}{c}{\textbf{2014}} & \multicolumn{3}{c}{\textbf{2015}} & \multicolumn{3}{c}{\textbf{2016}} \\ \midrule
\textbf{Attribute} & \textbf{\%Na} & \textbf{TVC} & \textbf{CA} & \textbf{\%Na} & \textbf{TVC} & \textbf{CA} & \textbf{\%Na} & \textbf{TVC} & \textbf{CA} \\ \midrule
Age & 0 & 1 & 1 & 0 & 1 & 1 & 0 & 1 & 1 \\
Sex & 0 & 1 & 1 & 0 & 1 & 1 & 0 & 1 & 1 \\
Work & 0 & 0.907 & 1 & 0.001 & 0.938 & 1 & 0.002 & 0.920 & 1 \\
DrvLic & 0.001 & 0.919 & 1 & 0 & 0.908 & 1 & 0 & 0.922 & 1 \\
EduLevel & 0.444 & 0.782 & 1 & 0.937 & 0.130 & 0.130 & 0.942 & 0.156 & 0.156 \\
LifeCatg & 0.555 & 0.931 & 1 & 0.062 & 0.905 & 1 & 0.056 & 0.924 & 1 \\ \bottomrule
\end{tabular}%
}
\end{table}

Subsequently, we performed a multi-dimensional evaluation for the joint distribution using the SRMSE across all six attributes ($k=6$ joint), representing all the attributes available in the dataset. This evaluation considers 4116 unique categorical combinations possible for the given attribute list. Table \ref{tab:srmse-travel} presents the number of unique categorical combinations for both the actual and synthetic data, along with the SRMSE score for each target year. The results indicate that the CT-GAN generates synthetic data with an acceptable average deviation of 5.346 unique combinations per bin across most target years, excluding 2015 and 2016. For these two years, a higher level of error is observed due to substantial missing data in the target dataset, as evidenced by the notably low number of unique combinations in the target data for 2015 and 2016.

\begin{table}[htbp]
\centering
\caption{Multi-dimension evaluation for joint distribution of six attributes with 4116 possible unique categorical combinations.}
\label{tab:srmse-travel}
\resizebox{0.5\columnwidth}{!}{%
\begin{tabular}{@{}cccc@{}}
\toprule
\textbf{Year} & \textbf{\begin{tabular}[c]{@{}c@{}}\# Comb.\\ Target\end{tabular}} & \textbf{\begin{tabular}[c]{@{}c@{}}\# Comb.\\ Synthetic\end{tabular}} & \textbf{SRMSE} \\ \midrule
2011 & 431 & 913 & 4.847 \\
2012 & 367 & 613 & 5.080 \\
2013 & 313 & 485 & 5.430 \\
2014 & 385 & 741 & 6.027 \\
2015 & 10 & 522 & 33.574 \\
2016 & 10 & 493 & 31.761 \\ \bottomrule
\end{tabular}%
}
\end{table}

\subsubsection{Evaluation on Zonal-level}\label{sec:ct-gan-zonal}
As described in Section \ref{Method}, one of this study's primary objectives involves analyzing performance improvements achievable by integrating CT-GAN with the traditional FBS-CO population synthesis framework. This section looks at and compares three models—the baseline (FBS-CO), stand-alone CT-GAN, and hybrid—using the Umeå-SAMS zonal-level dataset, based on the process shown in Figure \ref{fig:flowchart}. Mirroring the travel survey analysis, all models receive AGE and SEX marginal distributions as conditional variables per zone while treating WORK as an unconditional variable. Model performance is assessed through TVC metrics for all variables across predefined targets.

For all three models, we generate a synthetic zonal-level population utilizing the 2005 travel survey micro-sample as the base population. The baseline model (FBS-CO) and stand-alone model (CT-GAN) directly use the micro-sample to synthesize the target population. However, in the hybrid model, we combine FBS-CO with CT-GAN by initializing FBS-CO with the base population generated by CT-GAN. Specifically, we synthesize a base population using CT-GAN that is twice the size of the micro-sample, which is then fed to FBS-CO to synthesize the target population. 

Table \ref{tab:result_combined} presents the results for the synthetic population generation across all 89 Umeå-SAMS zones. The evaluation is based on the average RSSZ and average TVC scores for AGE, SEX, and WORK across all zones. For FBS-CO synthesis, a zone is considered successful when the RSSZ score comparing the target marginals and the synthetic population marginals is below 1. On the other hand, a zone is deemed successful if the CT-GAN generates the exact same total population as the target.

\begin{table}[htbp]
\centering
\caption{Result of the population generation for 89 SAMS zone in Umeå using three different models.}
\label{tab:result_combined}
\resizebox{\columnwidth}{!}{%
\begin{tabular}{@{}lccccc@{}}
\toprule
\multicolumn{1}{c}{\multirow{2}{*}{\textbf{Status}}} & \multirow{2}{*}{\textbf{\#Zones}} & \multirow{2}{*}{\textbf{Avg. RSSZ}} & \multicolumn{3}{c}{\textbf{Avg. TVC}} \\ \cmidrule(l){4-6} 
\multicolumn{1}{c}{} &  &  & \textbf{AGE} & \textbf{SEX} & \textbf{WORK} \\ \midrule
\multicolumn{6}{c}{\textbf{FBS-CO {[}Baseline{]}}} \\ \midrule
Successful & 62 & 0.767 $\pm$ 0.207 & 0.930 $\pm$ 0.110 & 0.985 $\pm$ 0.073 & 0.904 $\pm$ 0.126 \\
Un-successful & 22 & 10.746 $\pm$ 28.021 & 0.896 $\pm$ 0.050 & 0.965 $\pm$ 0.002 & 0.891 $\pm$ 0.082 \\
No Population & 5 & - & - & - & - \\ \midrule
\multicolumn{6}{c}{\textbf{CT-GAN {[}Stand-alone{]}}} \\ \midrule
Successful & 84 & - & 1.0 $\pm$ 0 & 1.0 $\pm$ 0 & 0.918 $\pm$ 0.138 \\
Un-successful & 0 & - & 0 & 0 & 0 \\
No Population & 5 & - & - & - & - \\ \midrule
\multicolumn{6}{c}{\textbf{CT-GAN + FBS-CO {[}Hybrid{]}}} \\ \midrule
Successful & 65 & 0.799 $\pm$ 0.201 & 0.957 $\pm$ 0.119 & 0.986 $\pm$ 0.057 & 0.918 $\pm$ 0.127 \\
Un-successful & 19 & 12.315 $\pm$ 30.671 & 0.901 $\pm$ 0.053 & 0.975 $\pm$ 0.001 & 0.907 $\pm$ 0.088 \\
No Population & 5 & - & - & - & - \\ \bottomrule
\end{tabular}%
}
\end{table}

From the results, it can be seen that the stand-alone CT-GAN model performs the best at synthesizing the target population for all 89 zones. The table clearly shows that the model meets the marginal targets perfectly for 84 (out of 89) zones, except in zones with no population (5 zones). This is evidenced by the perfect average TVC scores for AGE and SEX across all zones. The performance of the stand-alone CT-GAN is far better than both the baseline and hybrid models. For the baseline FBS-CO models, it was possible to synthesize the target population only for the 62 zones. Even among the converged zones, none produced a synthetic population that perfectly matched the given targets for AGE and SEX, evident from the average TVC scores, which are below 1 for both AGE and SEX. The hybrid model performs better than the baseline models, improving the number of converging zones from 62 to 65. Moreover, the average TVC values for all attributes improve compared to the population synthesized by the baseline model; however, they are still lower than the stand-alone model. 

\begin{figure}[htbp]
   \centering
   \begin{subfigure}{0.48\textwidth}
        \centering
        \includegraphics[scale=0.91]{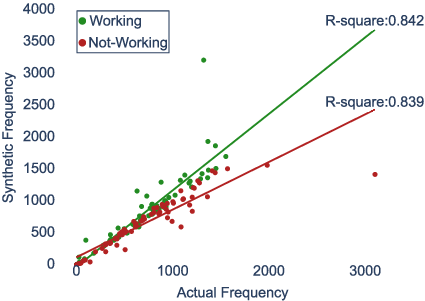}
        \caption{FBS-CO [Baseline]}
        \label{fig:baseline}
    \end{subfigure}
    \hfill
    \begin{subfigure}{0.48\textwidth}
        \centering
        \includegraphics[scale=0.91]{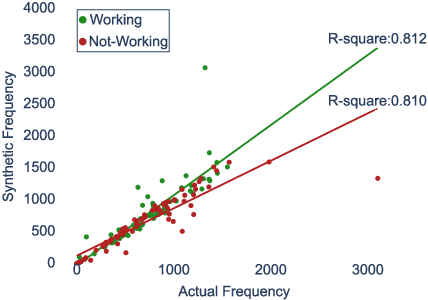}
        \caption{CT-GAN [Stand-alone]}
        \label{fig:standalone}
    \end{subfigure}
    
    \vspace{1em}  
    
    \begin{subfigure}{0.48\textwidth}
        \centering
        \includegraphics[scale=0.91]{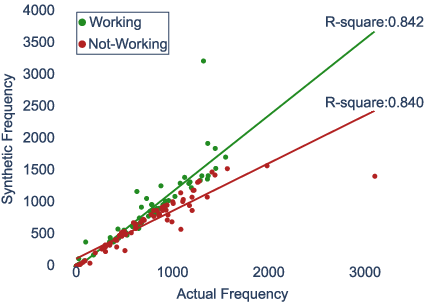}
        \caption{CT-GAN + FBS-CO [Hybrid]}
        \label{fig:hybrid}
    \end{subfigure}

    \caption{Scatter plot showing the frequency for WORK categories for both synthetic and actual population data.}
    \label{fig:work_scatter}
\end{figure}

Regarding the unconditional WORK attribute, all models exhibit comparable performance with minor variations. All the models were able to produce synthetic WORK data that closely approximates the ground truth, and this is evident from the TVC value of approximately 0.9 for zones that successfully converged. Even accounting for some error due to the conversion from 3-category to 2-category attributes, the models were able to generate synthetic populations that closely matched the target marginals for each zone. This is further illustrated in Figure \ref{fig:work_scatter}, showcasing the R-square calculated between the total number of working and not-working categories in each zone is as high as 0.81 for all models. Although the average TCV value for the stand-alone model is slightly lower than the other two models, the R-squared values indicate a similar level of performance across the models.

In summary, the results provide strong evidence of the stand-alone CT-GAN model's high performance in generating accurate target synthetic populations and performing better than the conventional FBS-CO. Even though the hybrid model has lower metrics compared to the stand-alone model, the results indicate that incorporating the CT-GAN-generated population as a base enhances FBS-CO by introducing more diverse combinations that are not present in the original micro-sample. Not only did it help in increasing the number of converged zones but also in improving the average TVC score across all attributes. 

\newpage
\section{Conclusion}\label{Conclusion}
In this study, we demonstrate that deep generative models, CT-GAN, can be effectively adapted for target population synthesis under user-defined aggregated marginal constraints. The intended outcome was to develop a method that not only meets specific conditional requirements but also preserves the inherent multi-dimensional interdependencies of the actual population.

Evaluation of both the travel survey data and the zonal-level population data, focusing on how well the synthetic data match the provided conditionals as well as the unconditioned attributes. Results reveal that CT-GAN reliably reproduces key demographic attributes. When tested on the travel survey dataset at the attribute level, the model was able to generate synthetic populations that exactly matched the conditional marginals for variables such as AGE and SEX across multiple target years. Furthermore, for the same dataset, for unconditioned attributes like DRVLIC, LIFECATG, and WORK, the synthetic data closely approximated the actual distributions, as indicated by high TVC and CA scores. The multi-dimensional evaluation further highlights CT-GAN’s capability, with the reasonable SRMSE values demonstrating an acceptable level of deviation in joint attribute distributions, except in cases where missing data substantially affected the target data quality. 

Additionally, the zonal-level assessment confirmed that the stand-alone CT-GAN model can synthesize populations that adhere closely to the given marginal constraints, achieving near-perfect TVC scores for conditioned attributes and exhibiting a strong correlation (R-squared of 0.81) between synthetic and actual WORK category frequencies.

We also analyzed the generative model using zonal data in a hybrid approach that combined CT-GAN with the FBS-CO method, comparing this hybrid with two stand-alone models: FBS-CO and CT-GAN. The results clearly indicate that a stand-alone CT-GAN performs best when compared with the alternatives. However, initializing FBS-CO with a CT-GAN-generated base population not only increased the number of converging zones but also improved overall marginal accuracy. This improvement demonstrates that providing a more descriptive base population to FBS-CO, containing out-of-training samples, enhances the overall performance of the optimization process. Moreover, this study highlights the ease of integrating generative models with traditional methods for population synthesis. This combination illustrates that merging deep generative models with traditional combinatorial optimization techniques can effectively mitigate limitations inherent to each approach when used independently.

In summary, this study confirms that CT-GAN is an effective method for target population synthesis, offering a robust framework for generating synthetic populations that respect both conditional constraints and complex joint distributions. The insights derived from our evaluations provide a solid foundation for future research, particularly in refining hybrid approaches that leverage the strengths of both statistical learning and deterministic optimization methods to overcome challenges related to data sparsity and imbalances.

\section{Future works}
A key limitation of the current work is that it only accounts for individual-level constraints in the population synthesis process. In reality, choices at the individual level are often made subjected to the influence of other household members. As such, it is important to have a good representation and reproduction of household structures to be able to realistically simulate intra-household interactions. Hence, the future work needs to address this need to incorporate both individual- and household-level constraints in the population synthesis process. A promising avenue for future research is how to generalize CT-GAN to perform multi-level population synthesis, such that data produced by it is consistent with both person-level and household-level constraints simultaneously.

A second weakness is the narrow range of constraints incorporated in the synthesis. In this study, we have limited the input constraints to only AGE and SEX, largely due to limitations in the data that was available for testing and validation. However, in real-world applications, population synthesis often involves a larger and more complex set of interdependent constraints. As the number and interdependence of such constraints grow, so does the difficulty of generating valid synthetic populations. Future research should therefore involve testing on larger and more diverse datasets, with more intricate numbers of entangled constraints. Because CT-GAN captures the underlying probabilistic distributions, it should be capable of handling high-dimensional constraint spaces more effectively than traditional SR methods. This expectation has to be rigorously empirically verified in future research.

Furthermore, while we can see that CT-GAN performs well, and promises to provide a worthwhile addition to the population synthesis toolbox, it does not by itself solve the problem of how to construct scenarios for future populations. The challenge is that we do neither know that the future population will have the same multidimensional distribution, nor what marginal distributions each zone will have.

A possible future extension is to see if a similar conditional generative method could be used to help with creating the marginal distributions. Machine learning is good at handling many dimensions in a way that planners are not. The planner most likely can handle a few key parameters they want to change for a future scenario, such as population growth, income change, or added housing by type. A model trained on all existing zones could then be used to fill in the rest of the marginals based on the conditions given by the planner.

\section{Acknowledgments}
The computations and data handling was enabled by the super-computing resource Berzelius provided by National Supercomputer Center at Linköping University and the Knut and Alice Wallenberg foundation. 

In the paper, we used AI tools, called Quillbot, specifically for grammar correction, language editing and paraphrasing sections of the paper. The authors have reviewed this and take full responsibility for the content of this paper.

\section{Author Contributions}
All the authors contributed to the study conception and methodology design. The design of the simulation study, data collection and analysis, and visualization were performed by Tanay Rastogi. The first draft of the manuscript was written by Tanay Rastogi, and all other authors reviewed and edited the latest versions of the manuscript. All authors read and approved the final manuscript.

\newpage
\bibliographystyle{apalike}
\bibliography{references}

\clearpage 
\newpage
\appendix

\section{Plot of loss for CT-GAN training}\label{apx:ct_gan_training}

\begin{figure}[ht]
\centering
{{\includegraphics[scale=0.57]{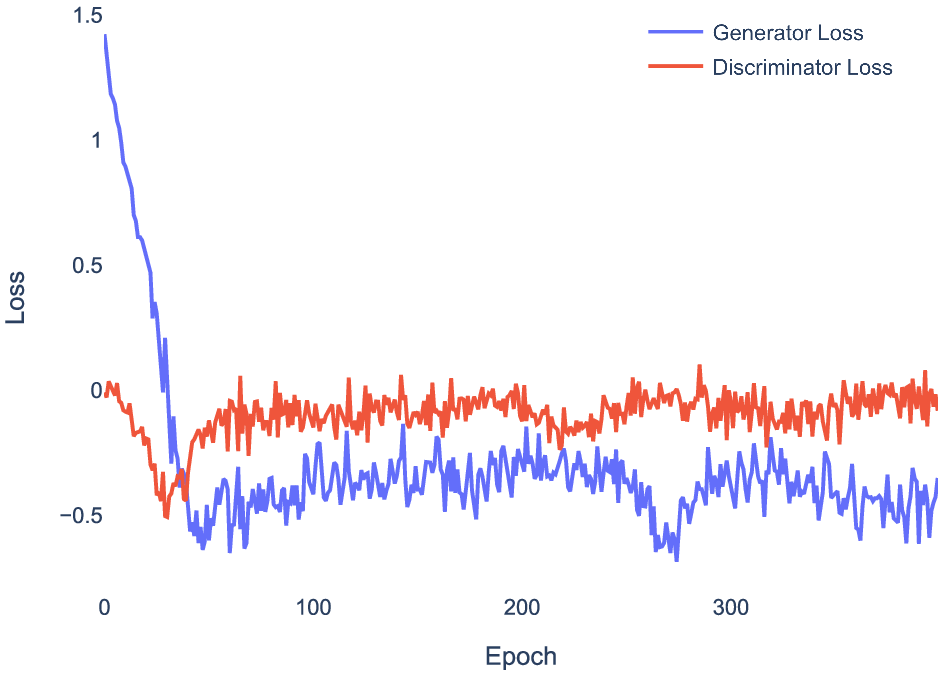}}}
\caption{Generator and Discriminator loss for CT-GAN while training on travel survey dataset.}
\end{figure}

\section{Bar graphs for travel surveys in different years}\label{apx:travel_survey_plots}

\begin{landscape}
\begin{figure}[htbp]
    \centering
    
    \begin{subfigure}{0.3\linewidth}
        \includegraphics[width=\linewidth]{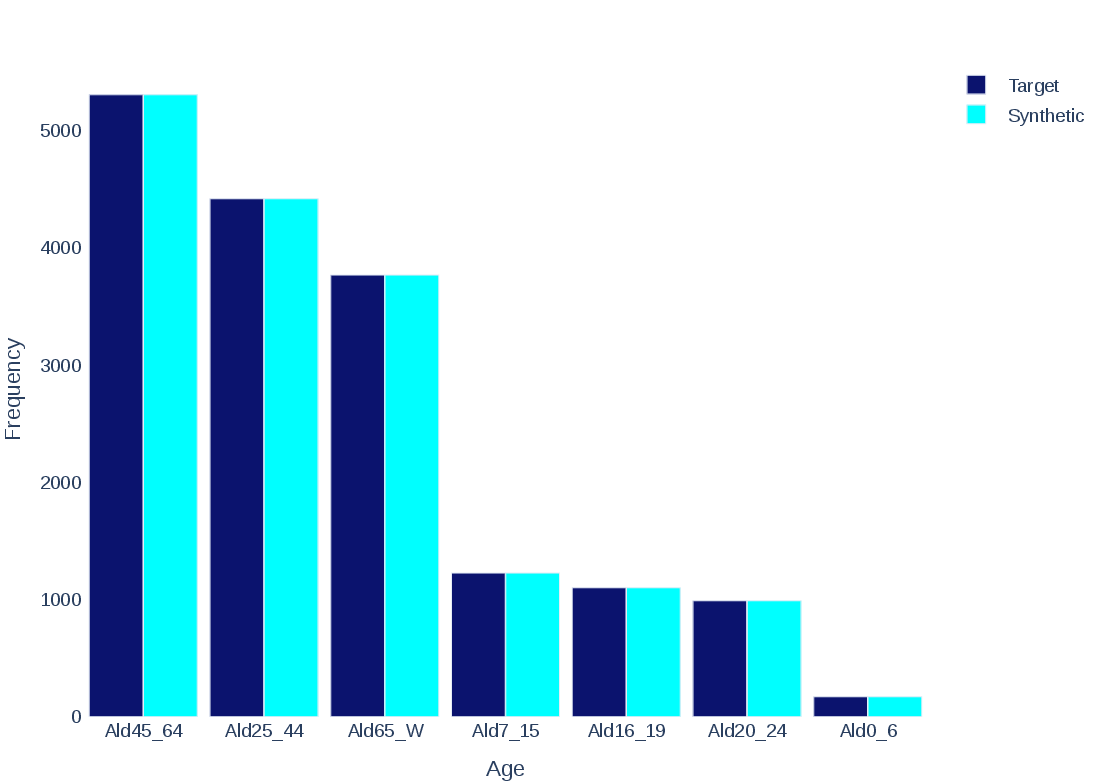}
        \caption{AGE}
    \end{subfigure}
    \hfill
    \begin{subfigure}{0.3\linewidth}
        \includegraphics[width=\linewidth]{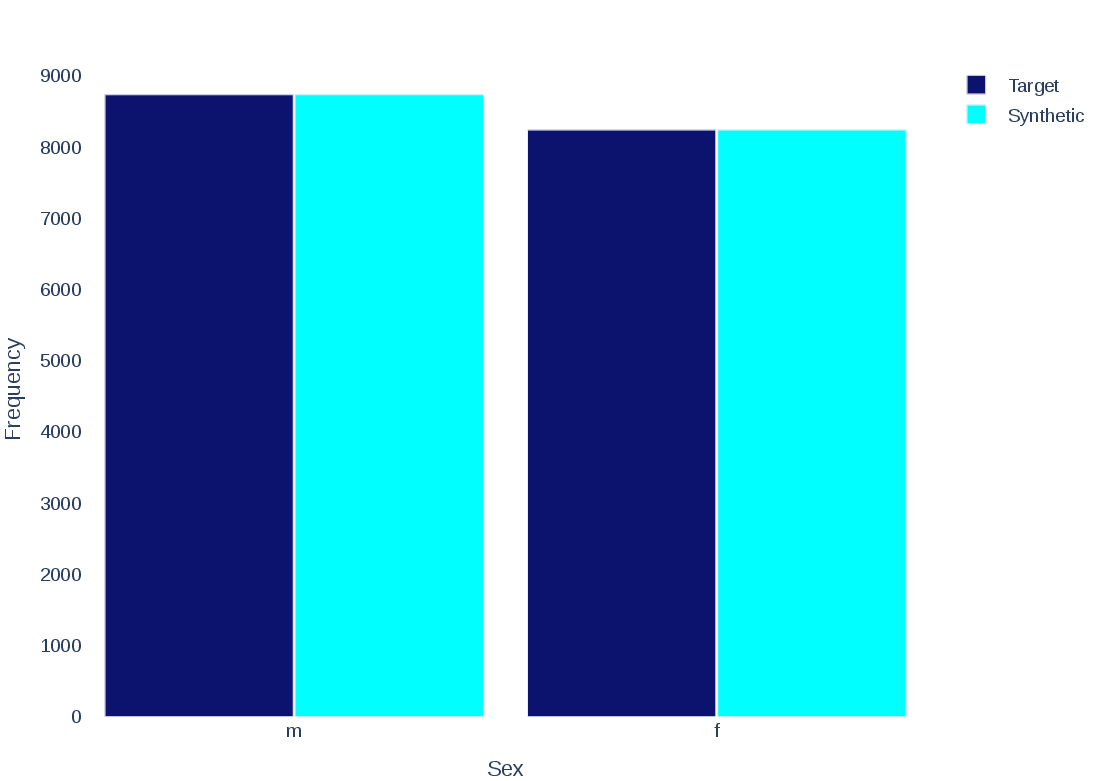}
        \caption{SEX}
    \end{subfigure}
    \hfill
    \begin{subfigure}{0.3\linewidth}
        \includegraphics[width=\linewidth]{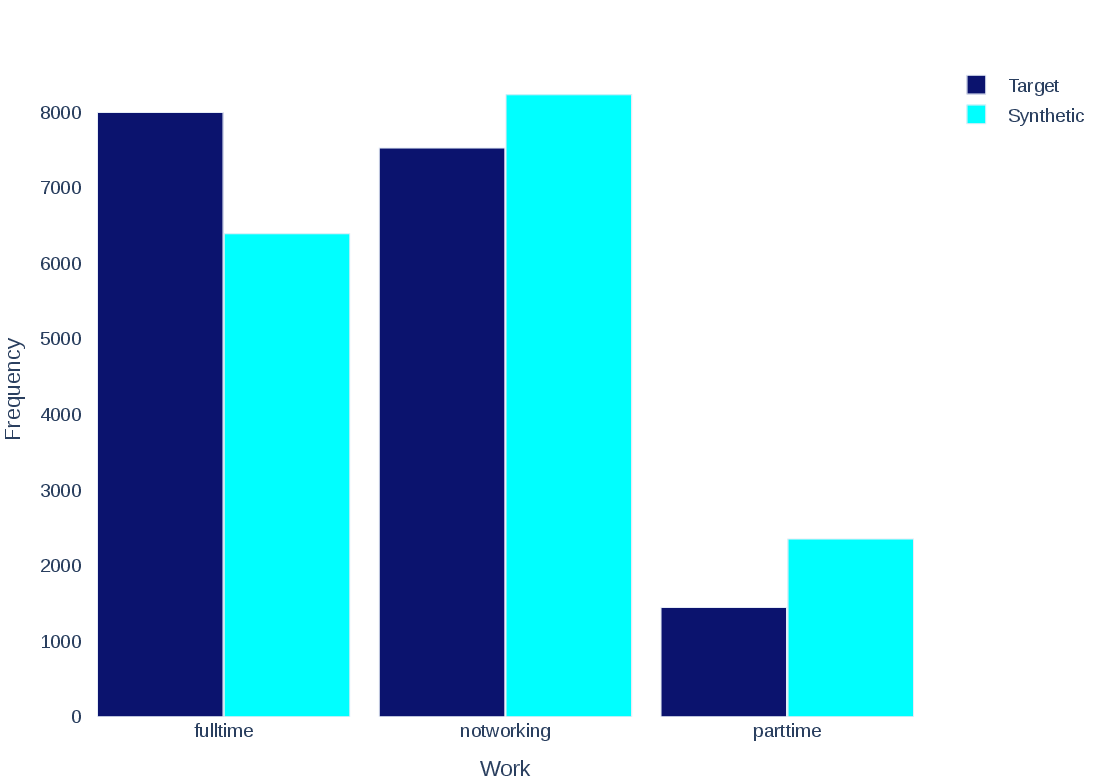}
        \caption{WORK}
    \end{subfigure}

    \vspace{1cm}

    \begin{subfigure}{0.3\linewidth}
        \includegraphics[width=\linewidth]{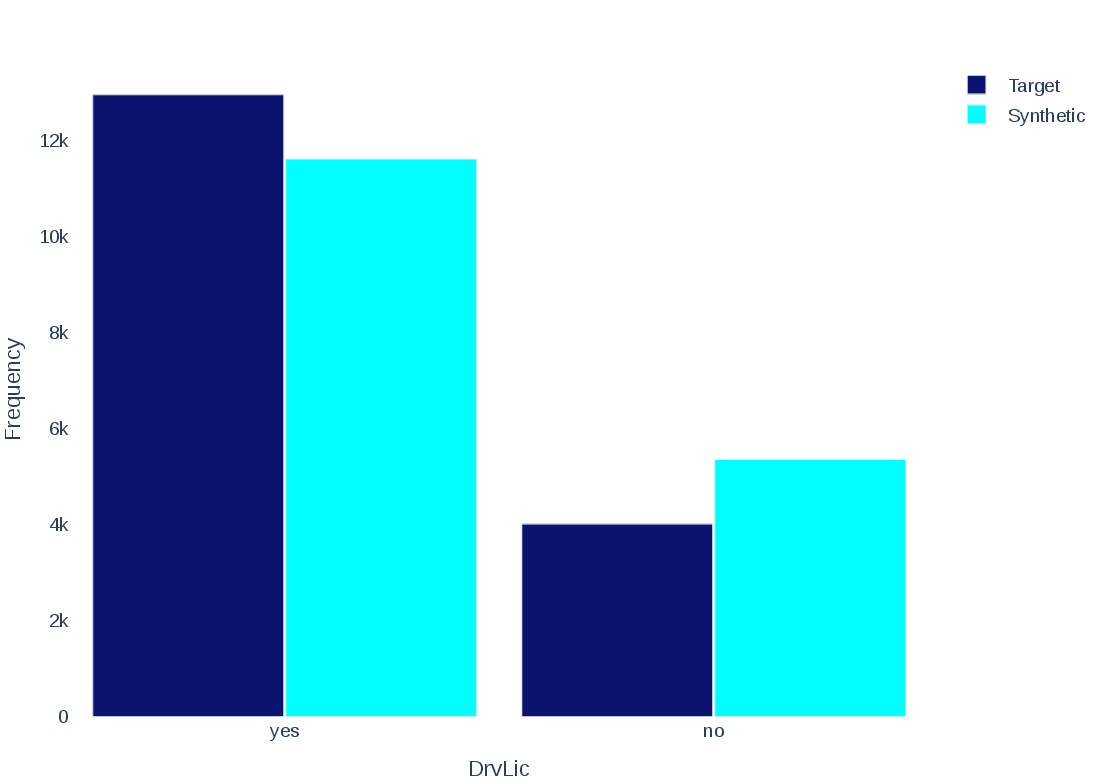}
        \caption{DRVLIC}
    \end{subfigure}
    \hfill
    \begin{subfigure}{0.3\linewidth}
        \includegraphics[width=\linewidth]{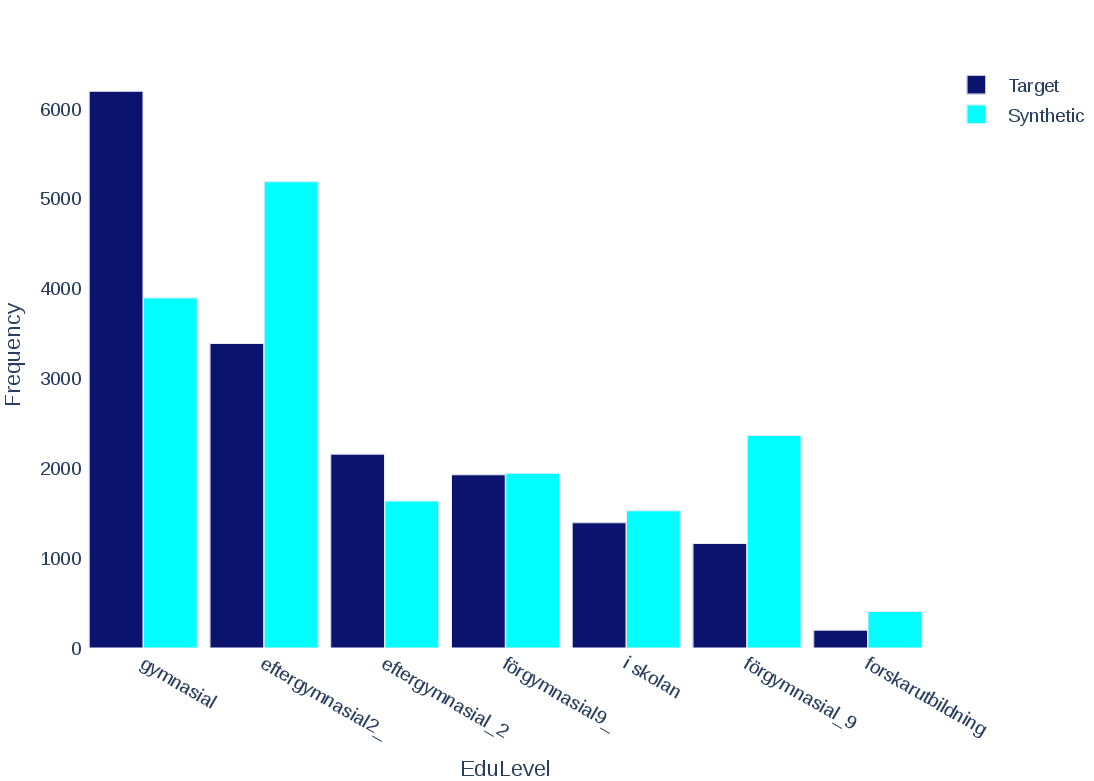}
        \caption{EDULEVEL}
    \end{subfigure}
    \hfill
    \begin{subfigure}{0.3\linewidth}
        \includegraphics[width=\linewidth]{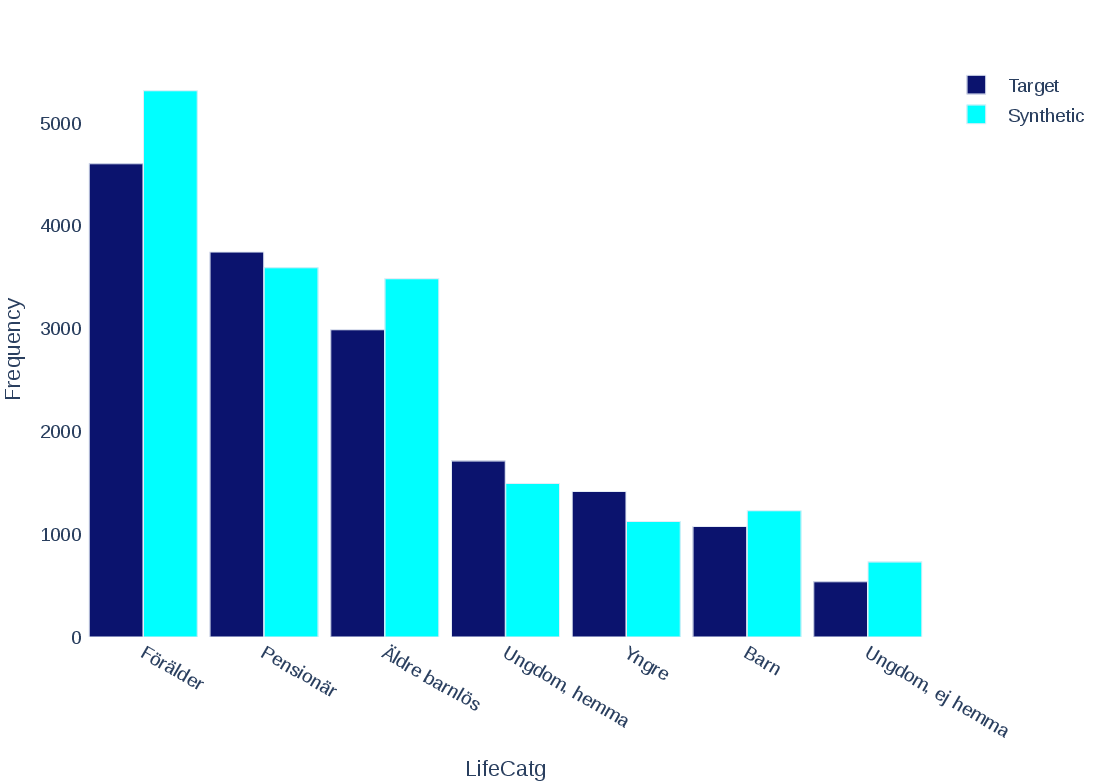}
        \caption{LIFECATG}
    \end{subfigure}

    \caption{Attribute distribution for target and synthetic travel survey data for year 2011.}

\end{figure}

\end{landscape}

\newpage
\begin{landscape}
\begin{figure}[htbp]
    \centering
    
    \begin{subfigure}{0.3\linewidth}
        \includegraphics[width=\linewidth]{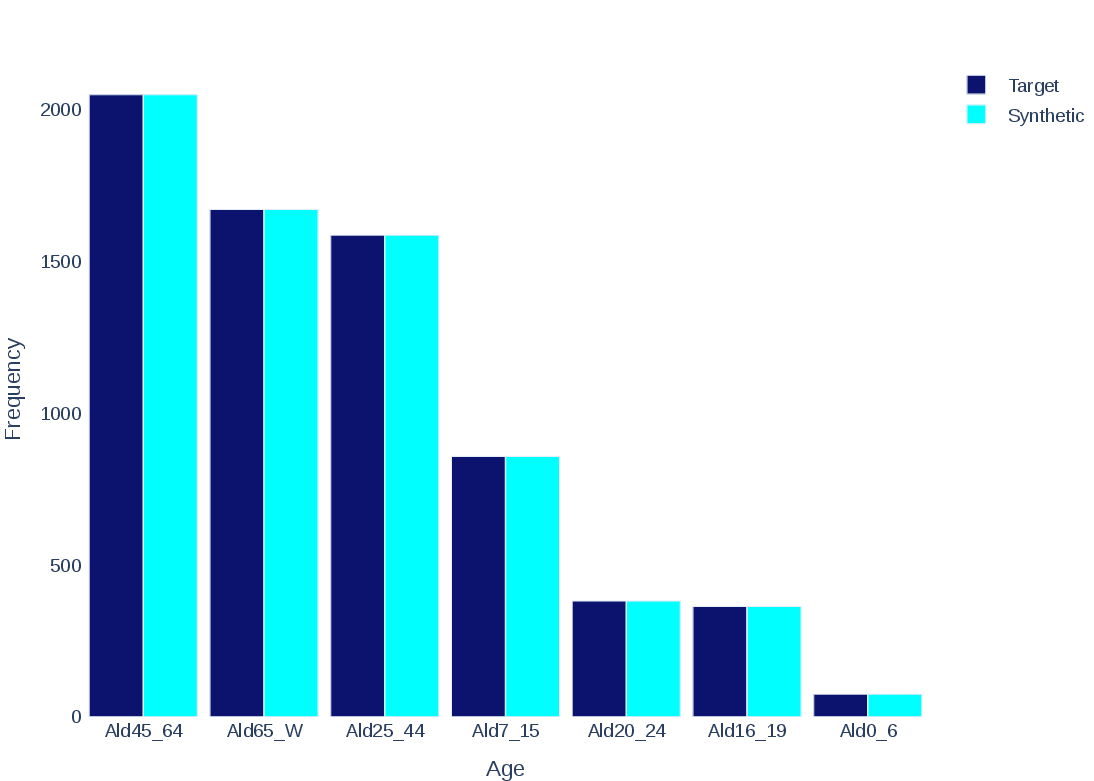}
        \caption{AGE}
    \end{subfigure}
    \hfill
    \begin{subfigure}{0.3\linewidth}
        \includegraphics[width=\linewidth]{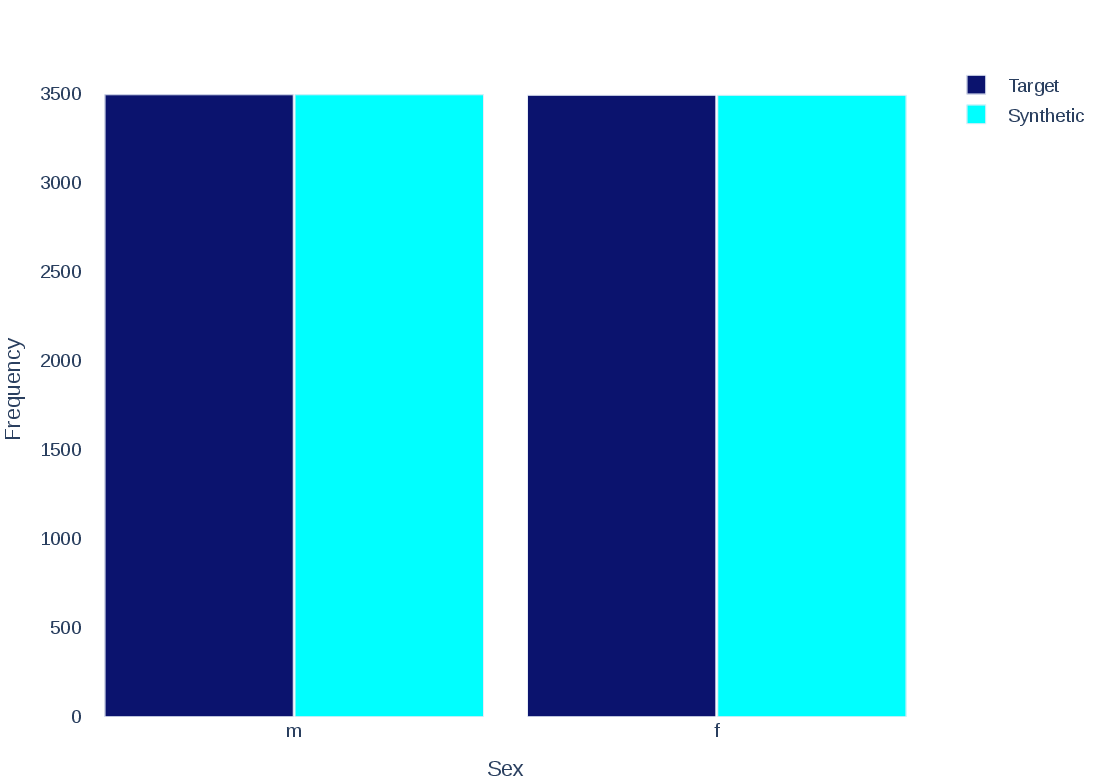}
        \caption{SEX}
    \end{subfigure}
    \hfill
    \begin{subfigure}{0.3\linewidth}
        \includegraphics[width=\linewidth]{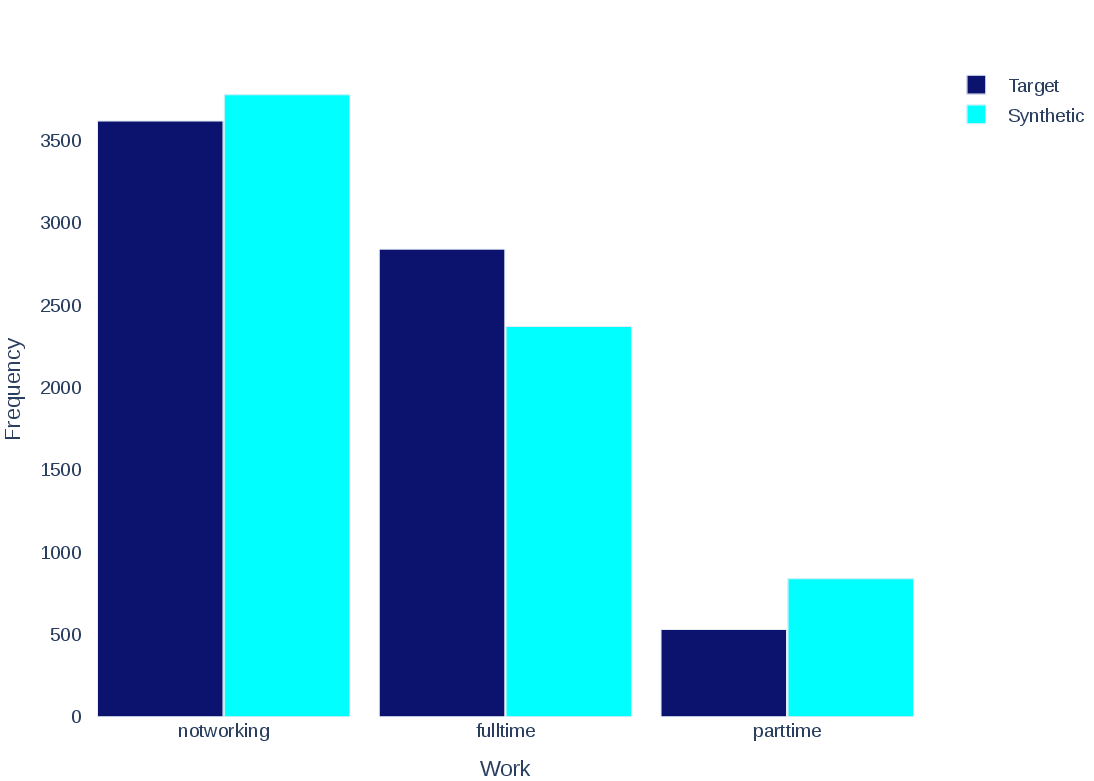}
        \caption{WORK}
    \end{subfigure}

    \vspace{1cm}

    \begin{subfigure}{0.3\linewidth}
        \includegraphics[width=\linewidth]{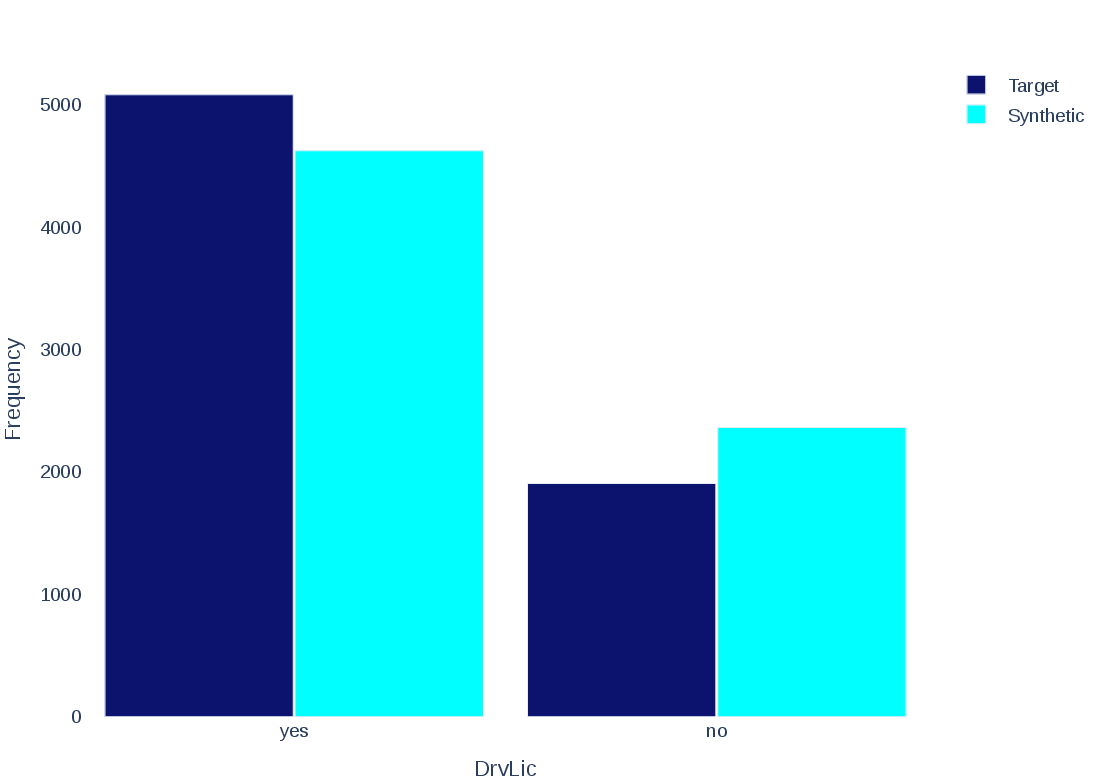}
        \caption{DRVLIC}
    \end{subfigure}
    \hfill
    \begin{subfigure}{0.3\linewidth}
        \includegraphics[width=\linewidth]{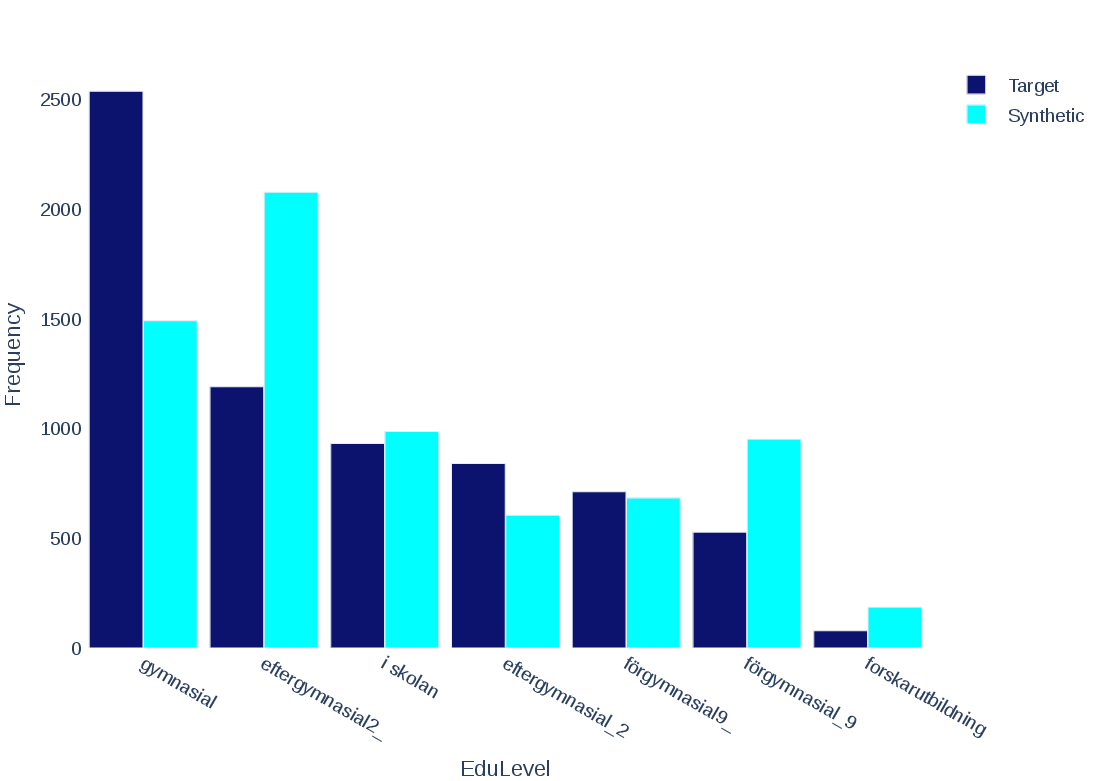}
        \caption{EDULEVEL}
    \end{subfigure}
    \hfill
    \begin{subfigure}{0.3\linewidth}
        \includegraphics[width=\linewidth]{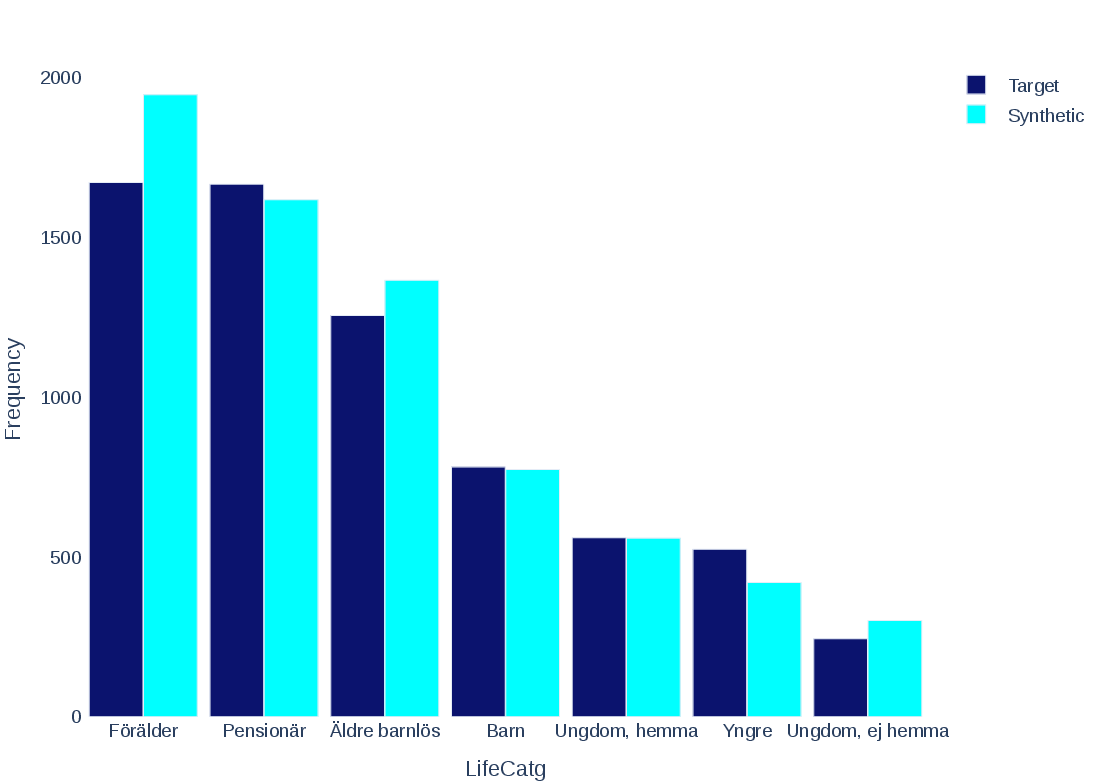}
        \caption{LIFECATG}
    \end{subfigure}

    \caption{Attribute distribution for target and synthetic travel survey data for year 2012.}

\end{figure}

\end{landscape}

\newpage
\begin{landscape}
\begin{figure}[htbp]
    \centering
    
    \begin{subfigure}{0.3\linewidth}
        \includegraphics[width=\linewidth]{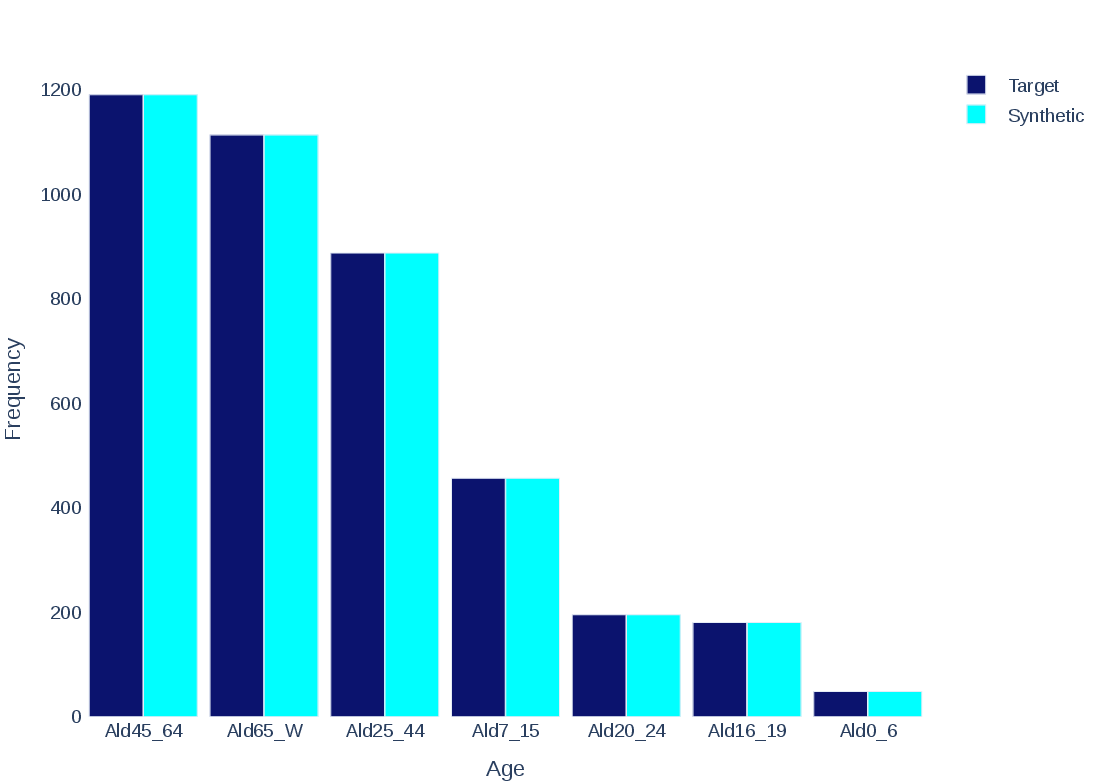}
        \caption{AGE}
    \end{subfigure}
    \hfill
    \begin{subfigure}{0.3\linewidth}
        \includegraphics[width=\linewidth]{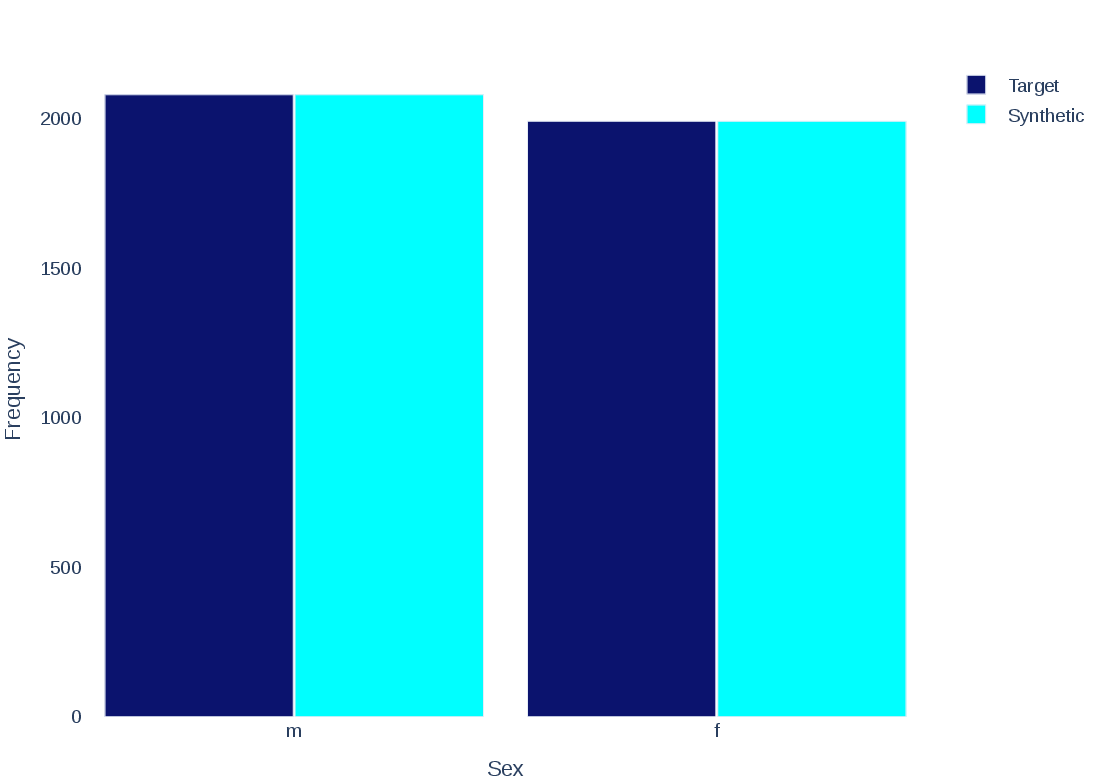}
        \caption{SEX}
    \end{subfigure}
    \hfill
    \begin{subfigure}{0.3\linewidth}
        \includegraphics[width=\linewidth]{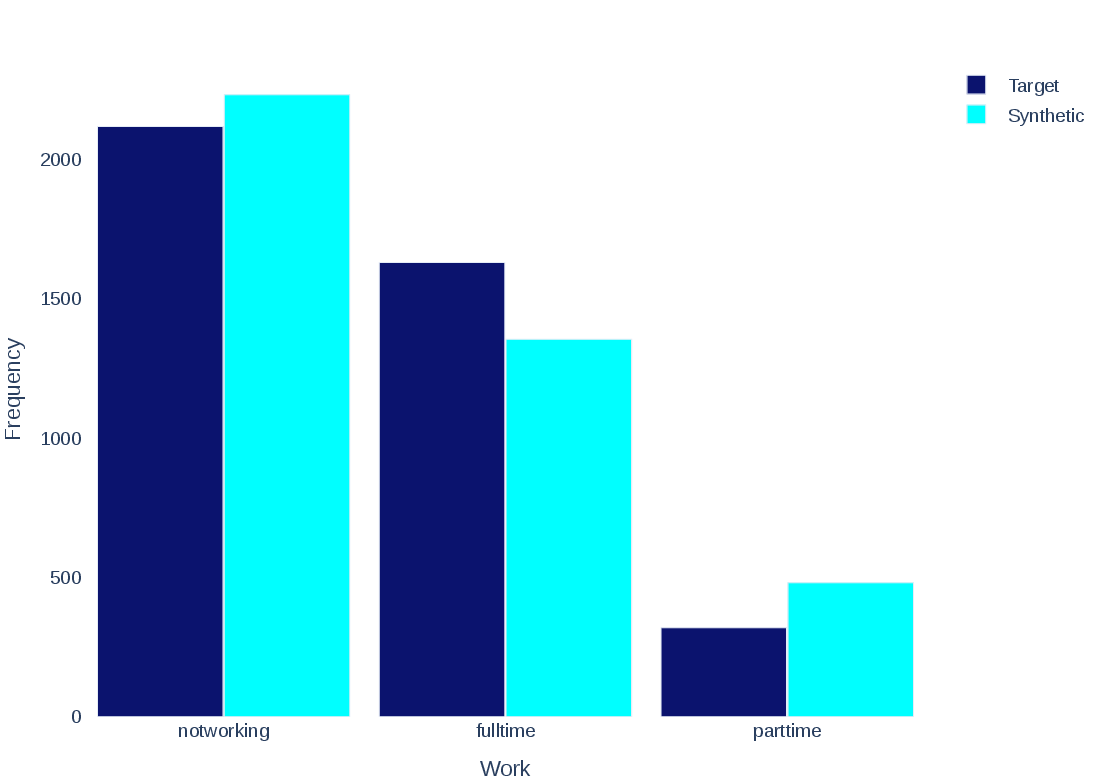}
        \caption{WORK}
    \end{subfigure}

    \vspace{1cm}

    \begin{subfigure}{0.3\linewidth}
        \includegraphics[width=\linewidth]{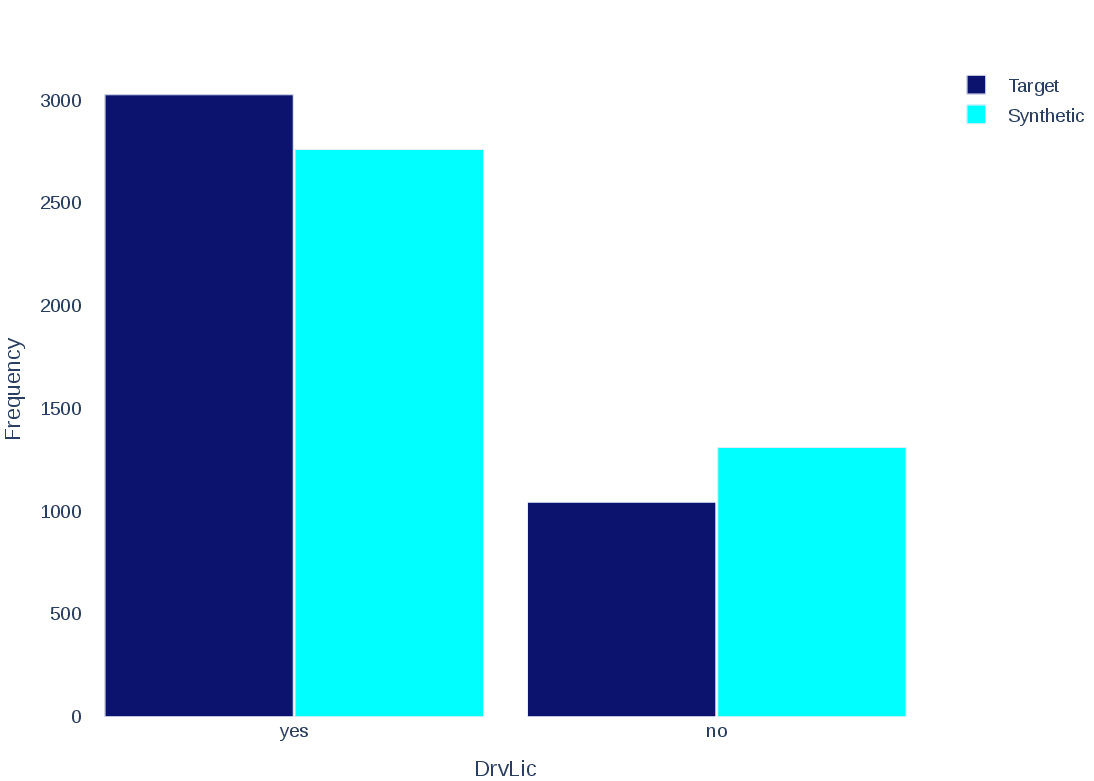}
        \caption{DRVLIC}
    \end{subfigure}
    \hfill
    \begin{subfigure}{0.3\linewidth}
        \includegraphics[width=\linewidth]{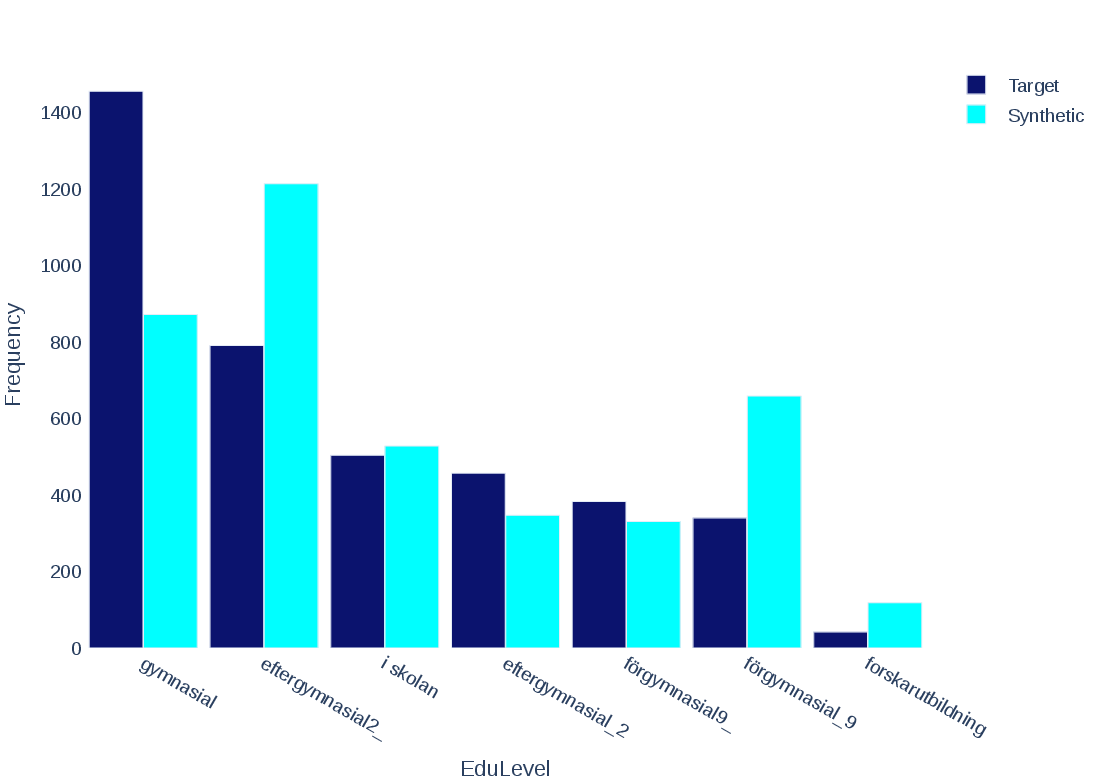}
        \caption{EDULEVEL}
    \end{subfigure}
    \hfill
    \begin{subfigure}{0.3\linewidth}
        \includegraphics[width=\linewidth]{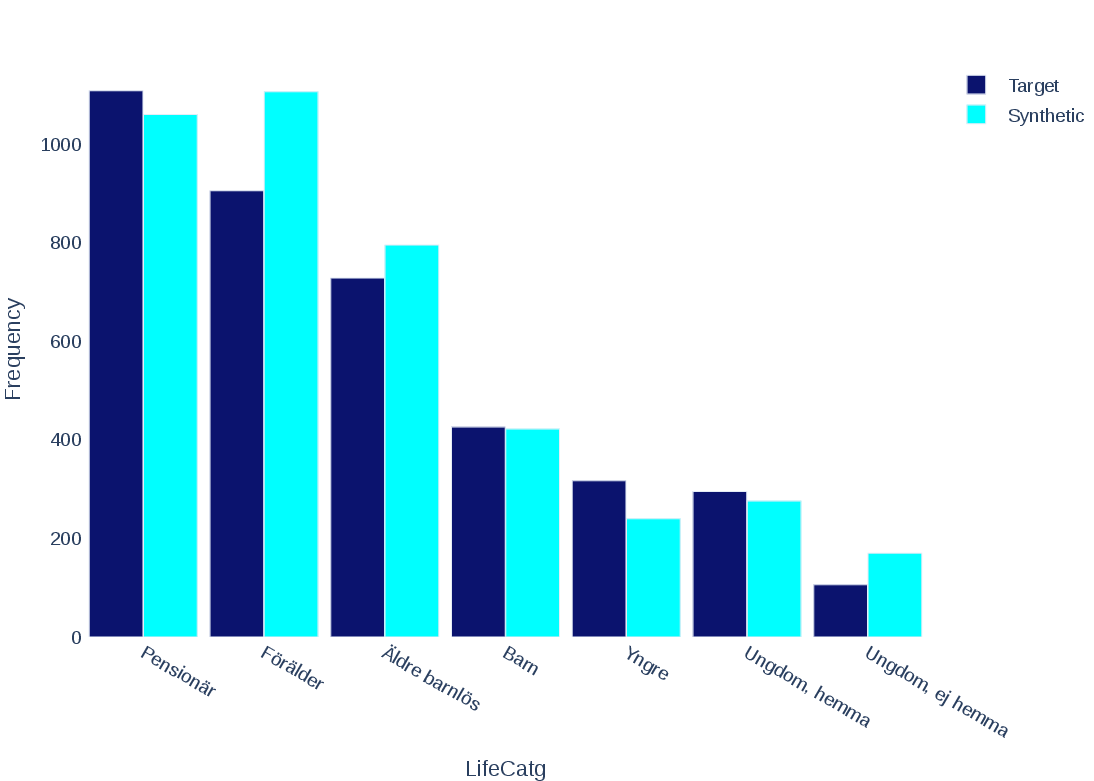}
        \caption{LIFECATG}
    \end{subfigure}

    \caption{Attribute distribution for target and synthetic travel survey data for year 2013.}

\end{figure}

\end{landscape}

\newpage
\begin{landscape}
\begin{figure}[htbp]
    \centering
    
    \begin{subfigure}{0.3\linewidth}
        \includegraphics[width=\linewidth]{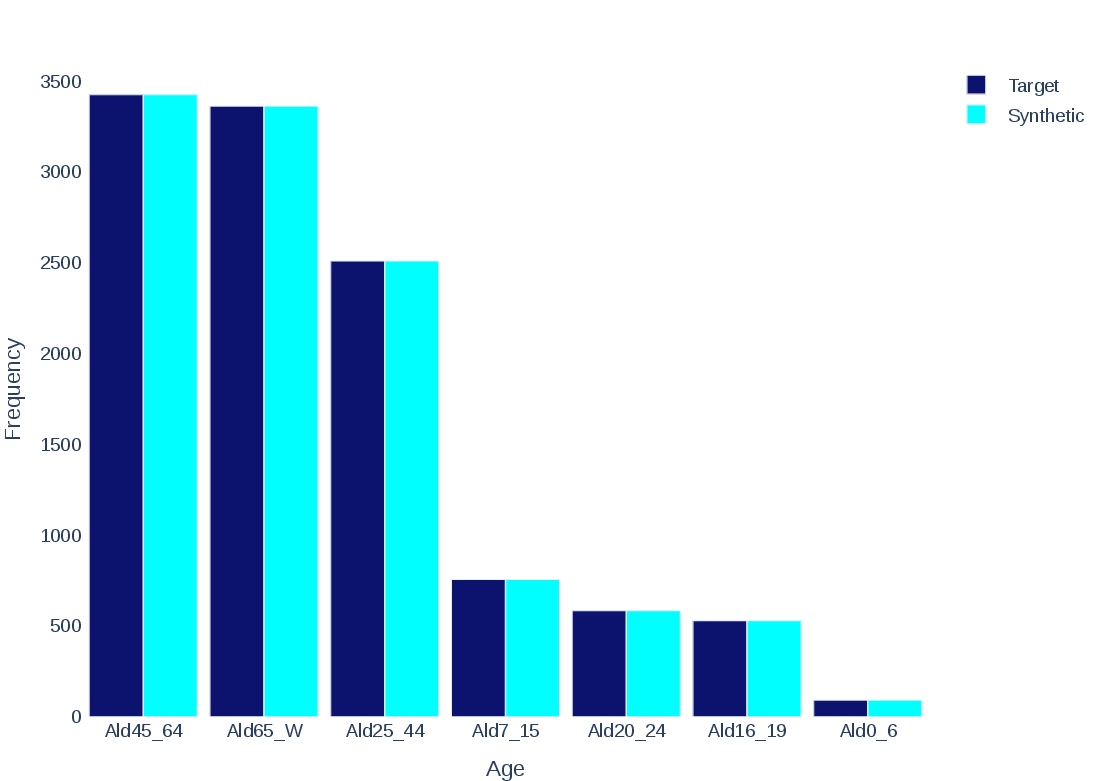}
        \caption{AGE}
    \end{subfigure}
    \hfill
    \begin{subfigure}{0.3\linewidth}
        \includegraphics[width=\linewidth]{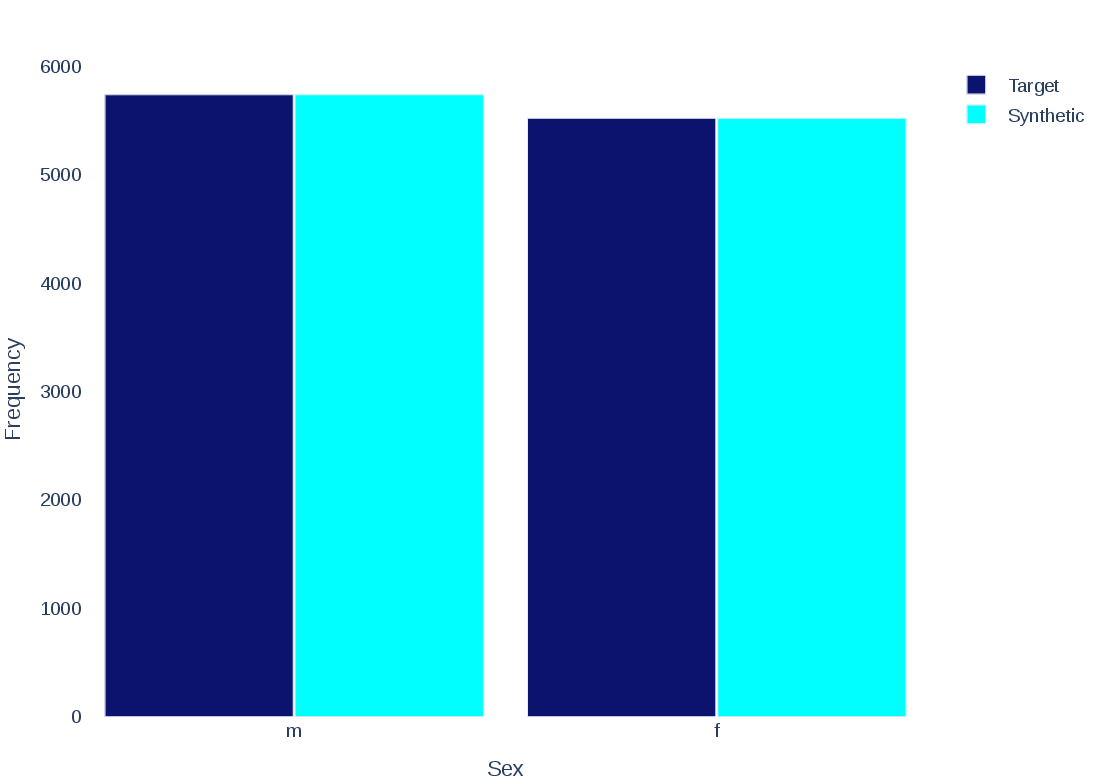}
        \caption{SEX}
    \end{subfigure}
    \hfill
    \begin{subfigure}{0.3\linewidth}
        \includegraphics[width=\linewidth]{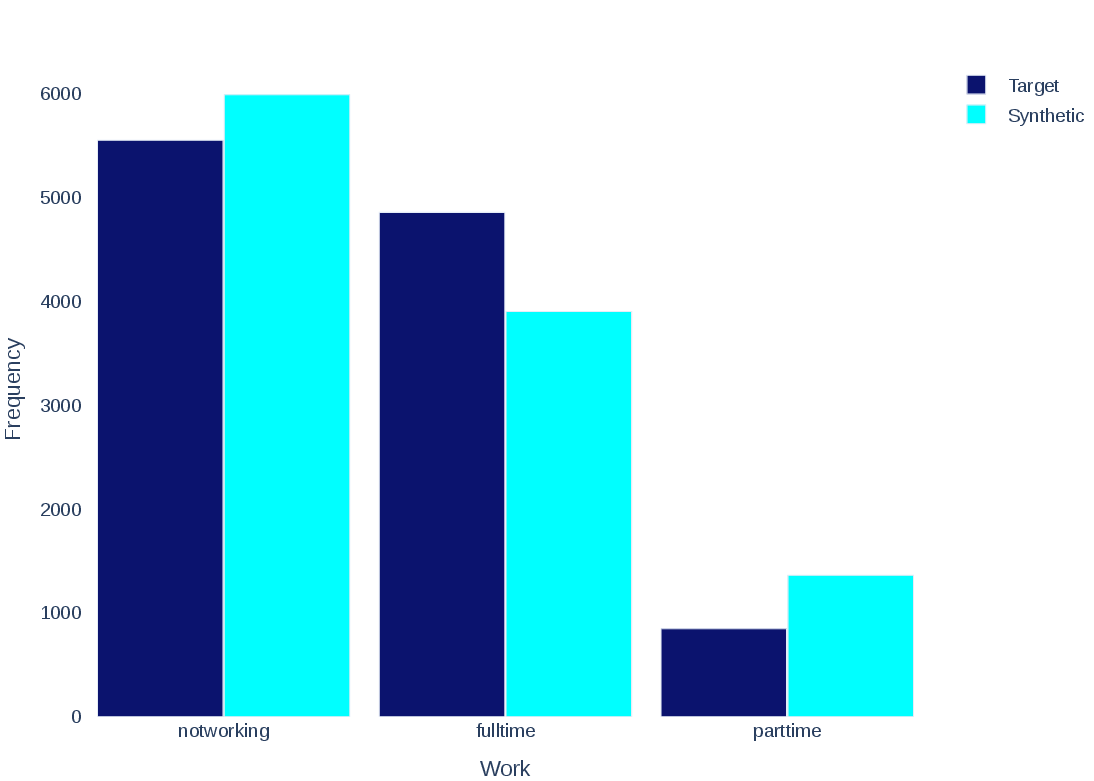}
        \caption{WORK}
    \end{subfigure}

    \vspace{1cm}

    \begin{subfigure}{0.3\linewidth}
        \includegraphics[width=\linewidth]{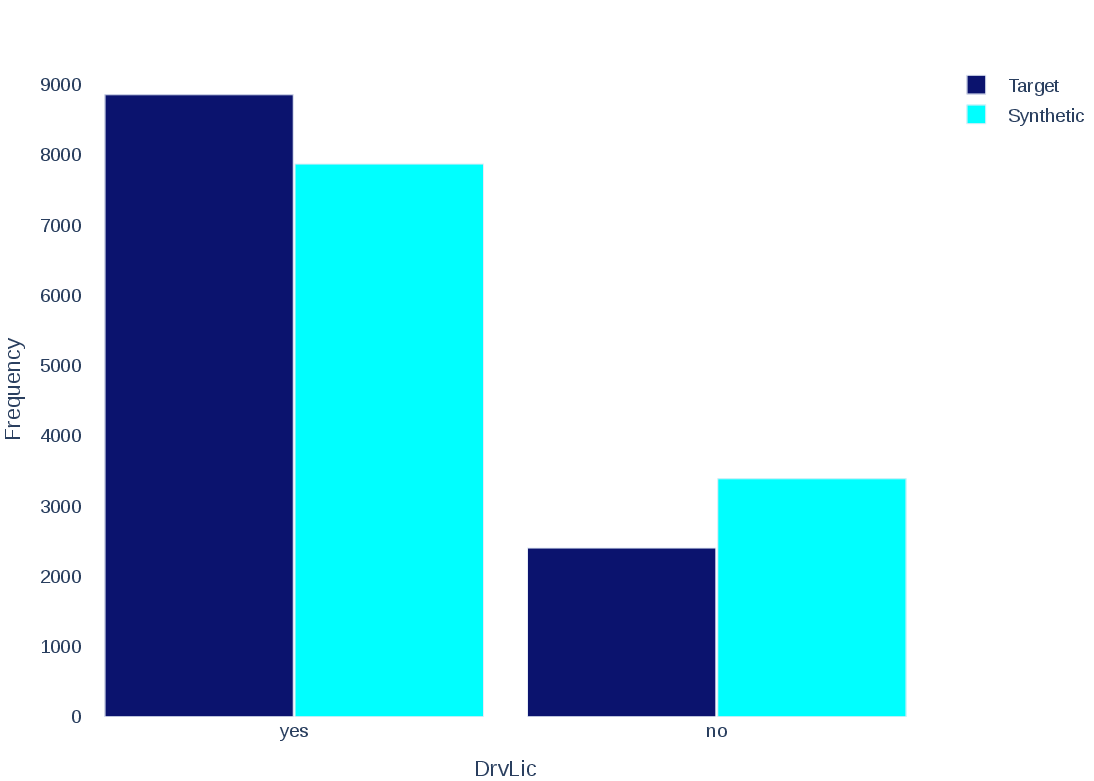}
        \caption{DRVLIC}
    \end{subfigure}
    \hfill
    \begin{subfigure}{0.3\linewidth}
        \includegraphics[width=\linewidth]{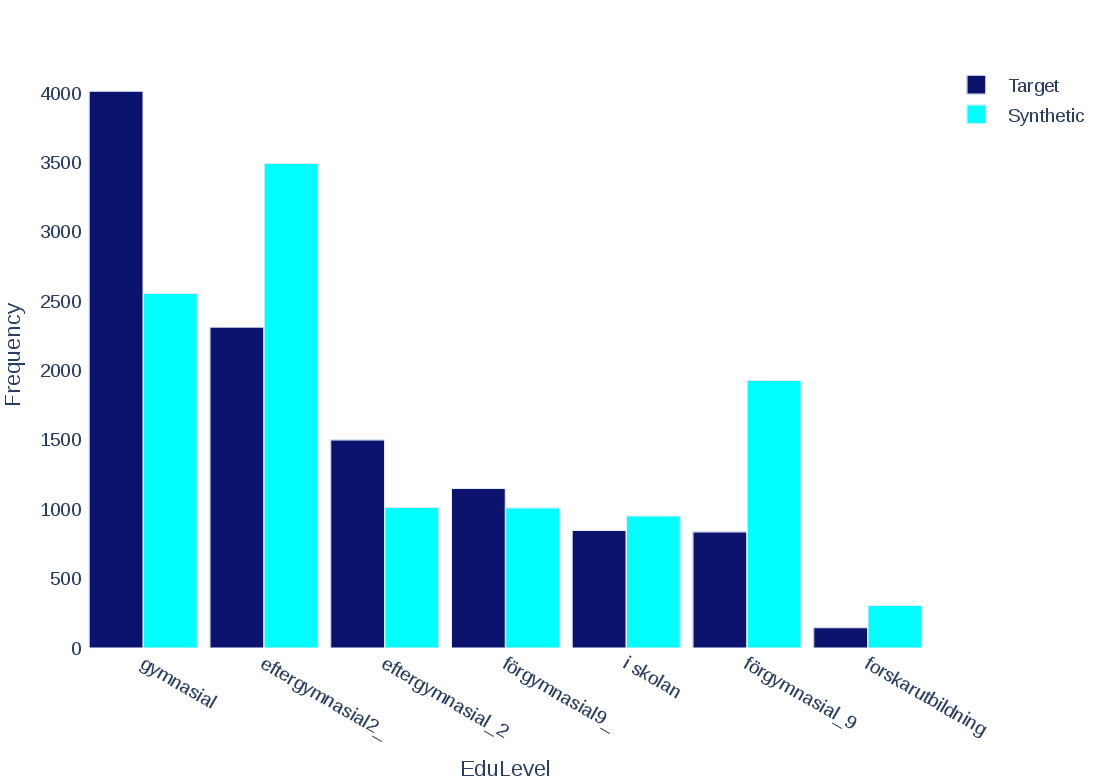}
        \caption{EDULEVEL}
    \end{subfigure}
    \hfill
    \begin{subfigure}{0.3\linewidth}
        \includegraphics[width=\linewidth]{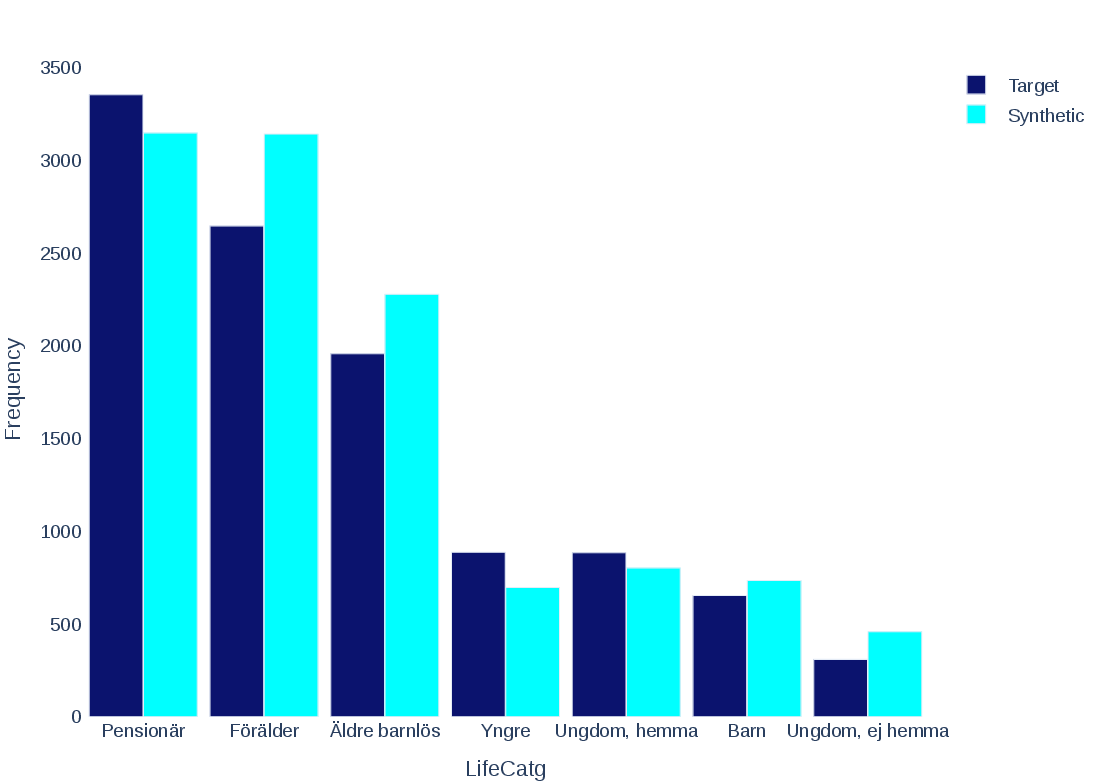}
        \caption{LIFECATG}
    \end{subfigure}

    \caption{Attribute distribution for target and synthetic travel survey data for year 2014.}

\end{figure}

\end{landscape}

\newpage
\begin{landscape}
\begin{figure}[htbp]
    \centering
    
    \begin{subfigure}{0.3\linewidth}
        \includegraphics[width=\linewidth]{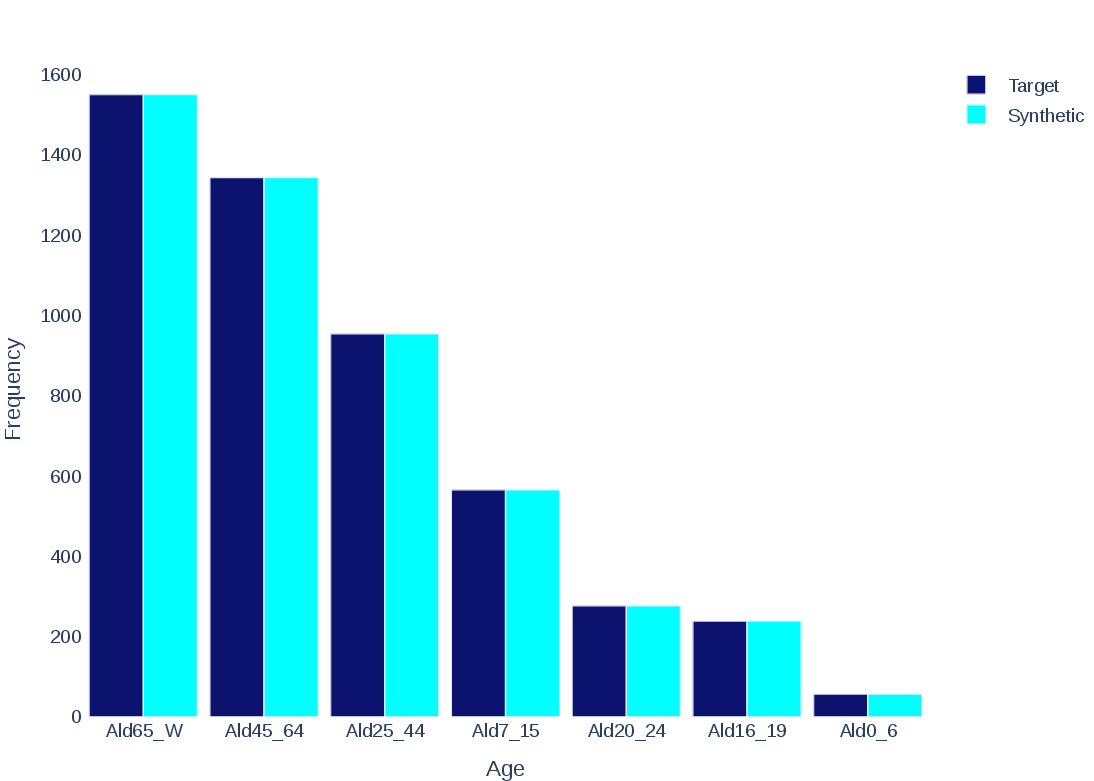}
        \caption{AGE}
    \end{subfigure}
    \hfill
    \begin{subfigure}{0.3\linewidth}
        \includegraphics[width=\linewidth]{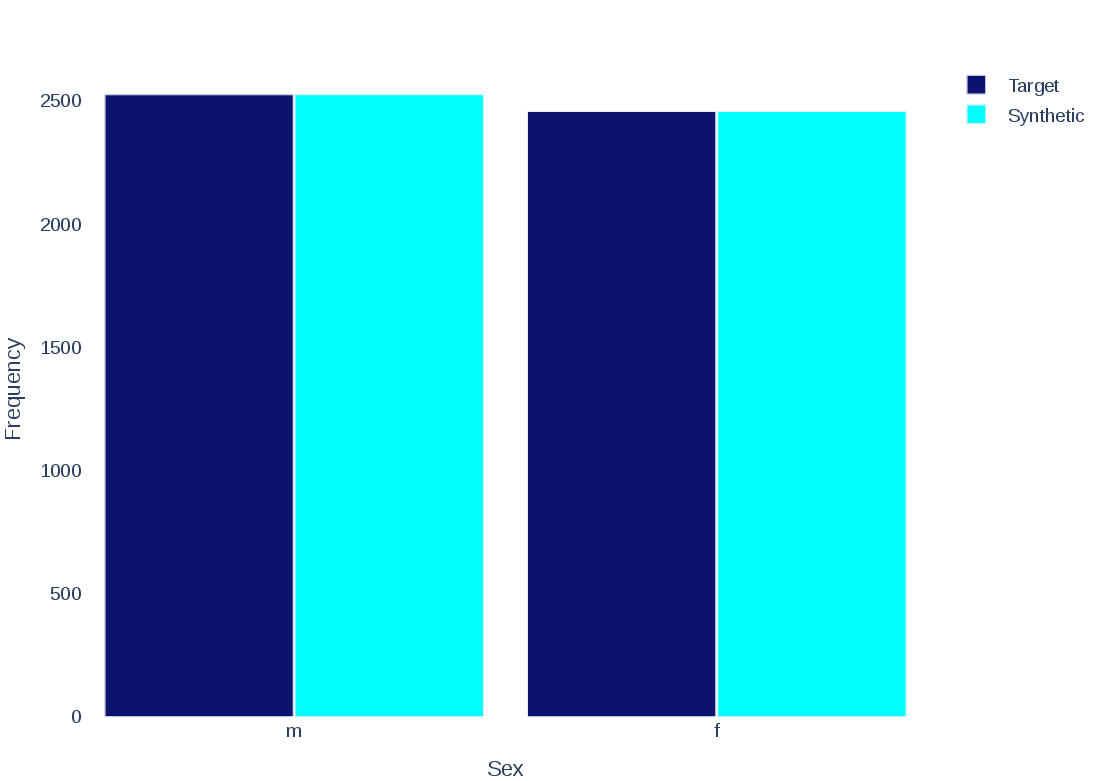}
        \caption{SEX}
    \end{subfigure}
    \hfill
    \begin{subfigure}{0.3\linewidth}
        \includegraphics[width=\linewidth]{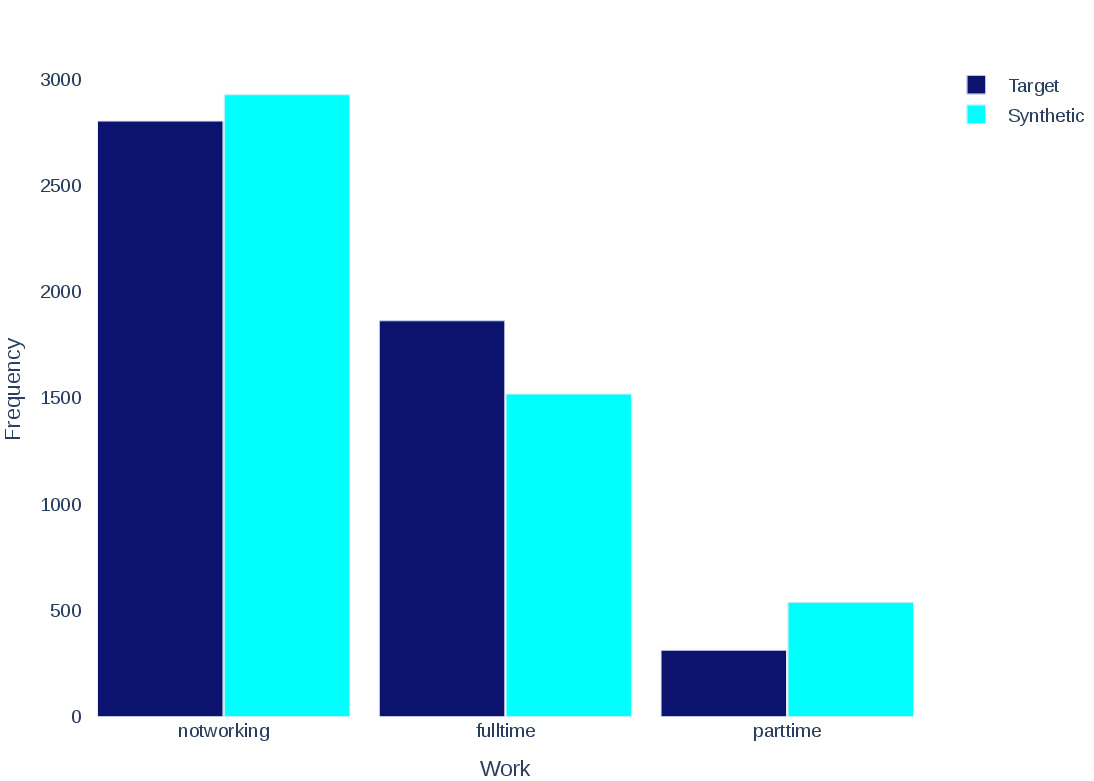}
        \caption{WORK}
    \end{subfigure}

    \vspace{1cm}

    \begin{subfigure}{0.3\linewidth}
        \includegraphics[width=\linewidth]{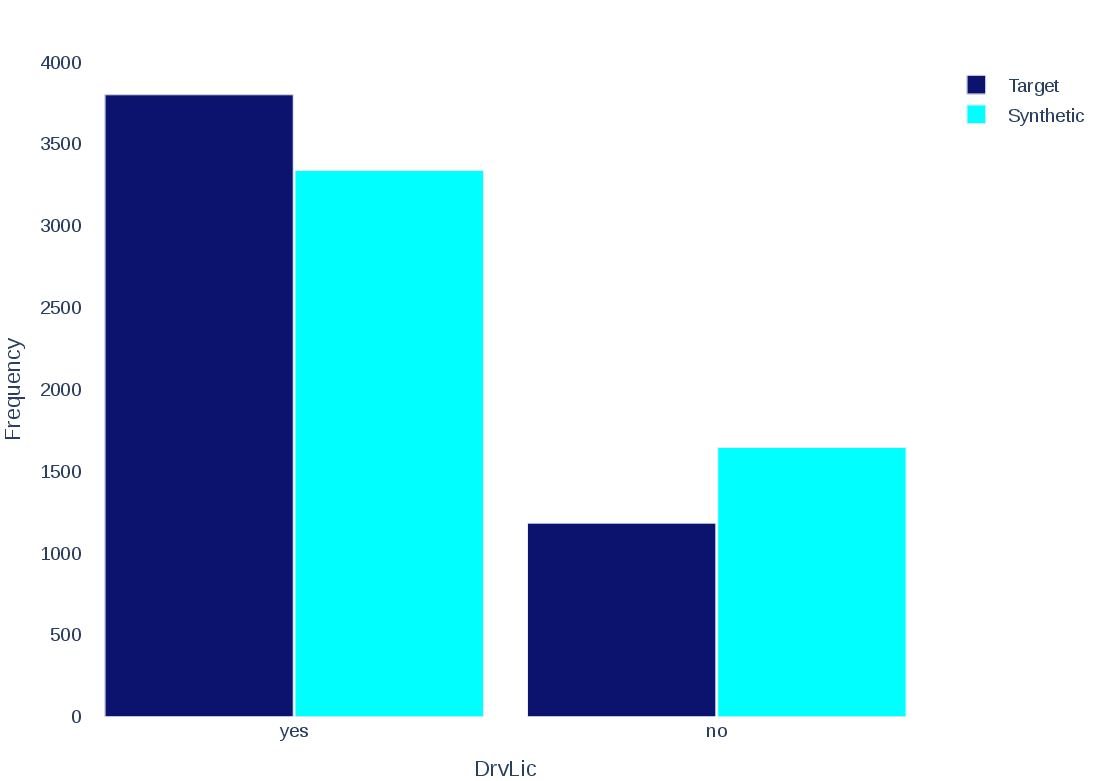}
        \caption{DRVLIC}
    \end{subfigure}
    \hfill
    \begin{subfigure}{0.3\linewidth}
        \includegraphics[width=\linewidth]{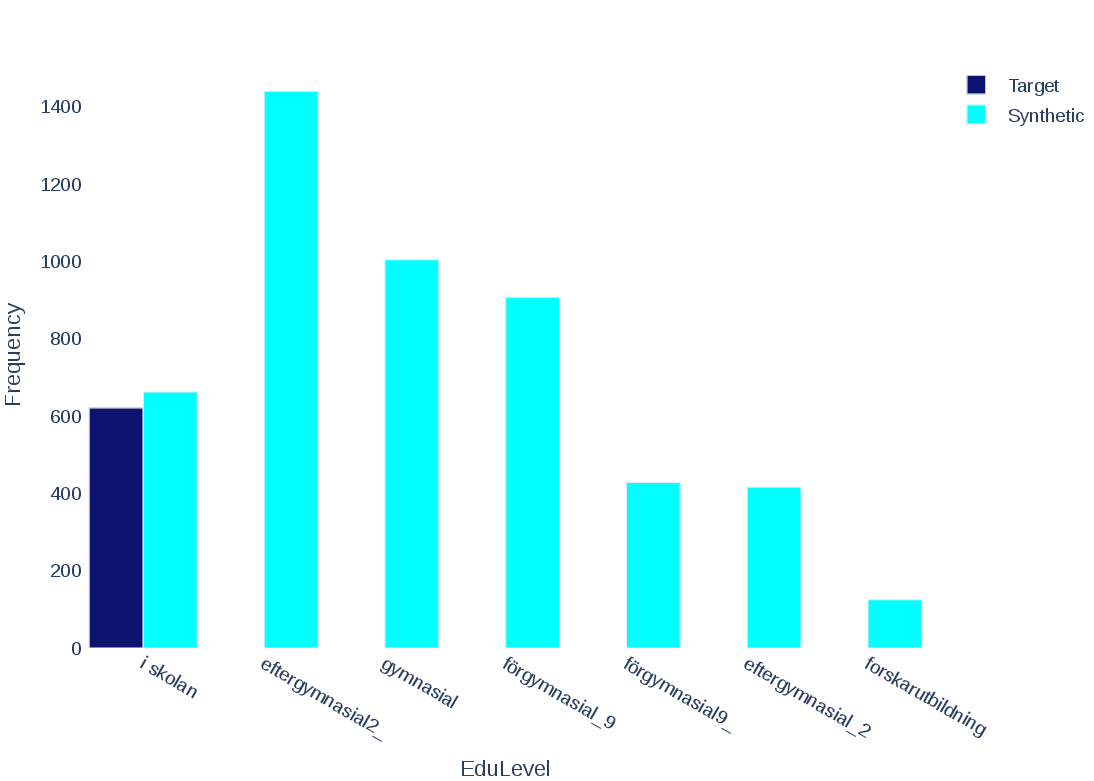}
        \caption{EDULEVEL}
    \end{subfigure}
    \hfill
    \begin{subfigure}{0.3\linewidth}
        \includegraphics[width=\linewidth]{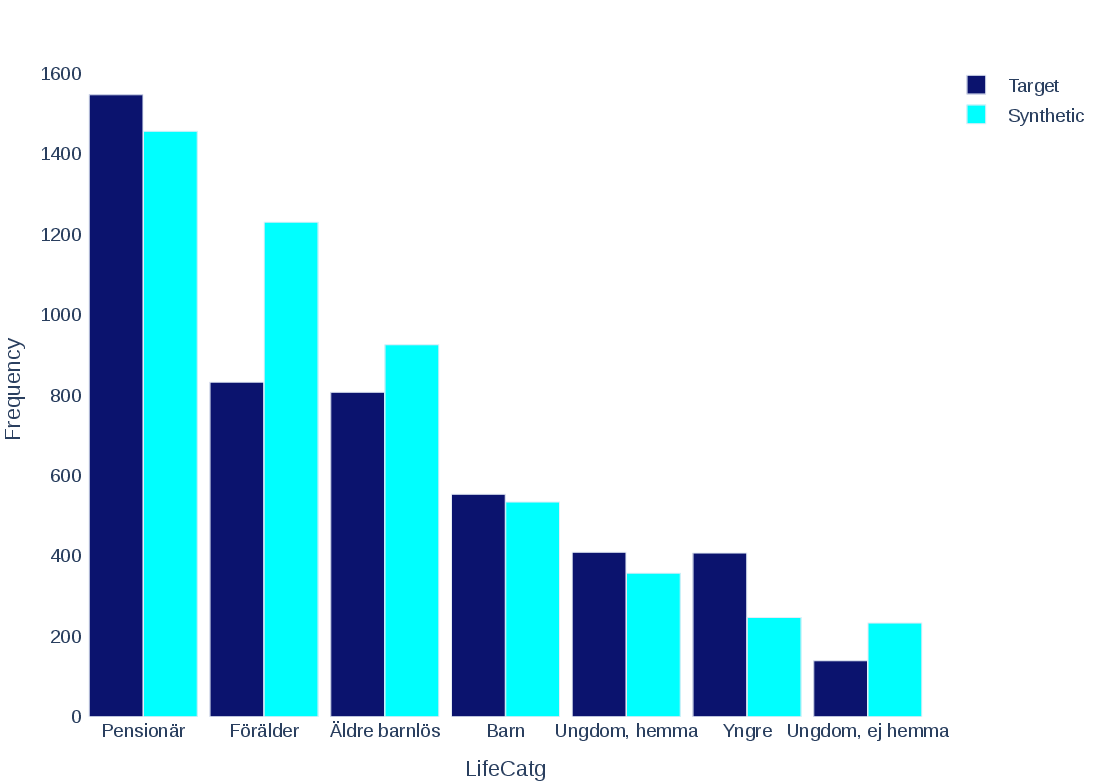}
        \caption{LIFECATG}
    \end{subfigure}

    \caption{Attribute distribution for target and synthetic travel survey data for year 2015.}

\end{figure}

\end{landscape}

\newpage
\begin{landscape}
\begin{figure}[htbp]
    \centering
    
    \begin{subfigure}{0.3\linewidth}
        \includegraphics[width=\linewidth]{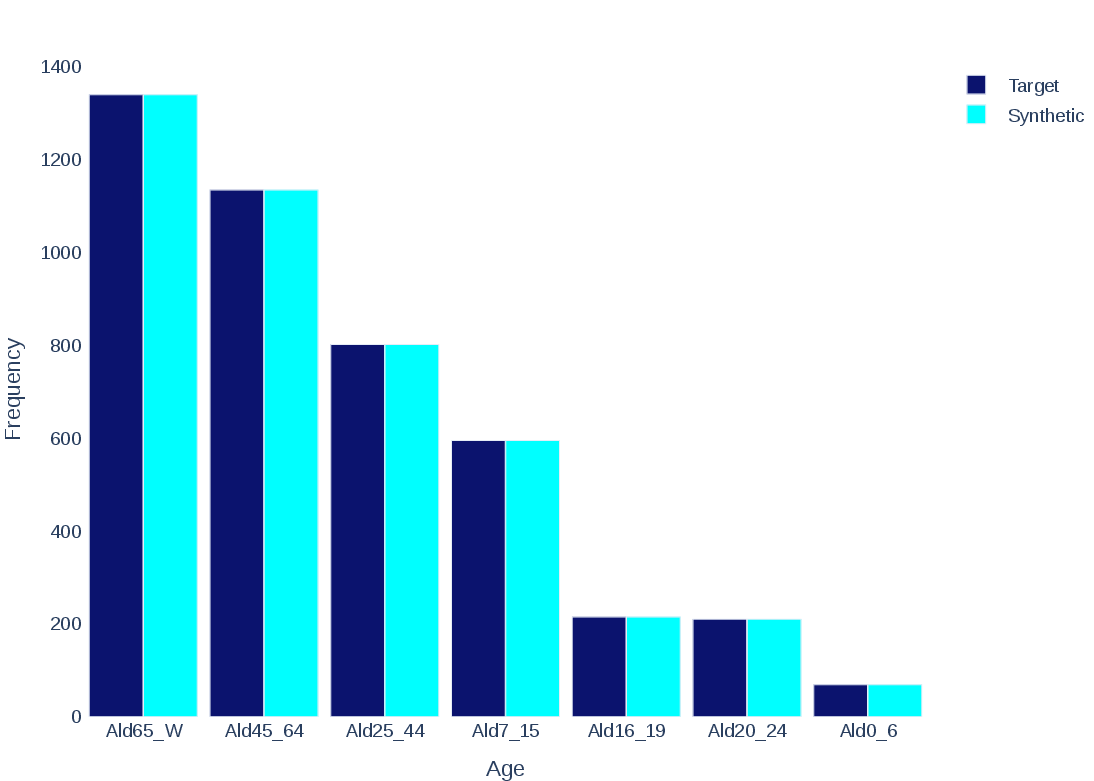}
        \caption{AGE}
    \end{subfigure}
    \hfill
    \begin{subfigure}{0.3\linewidth}
        \includegraphics[width=\linewidth]{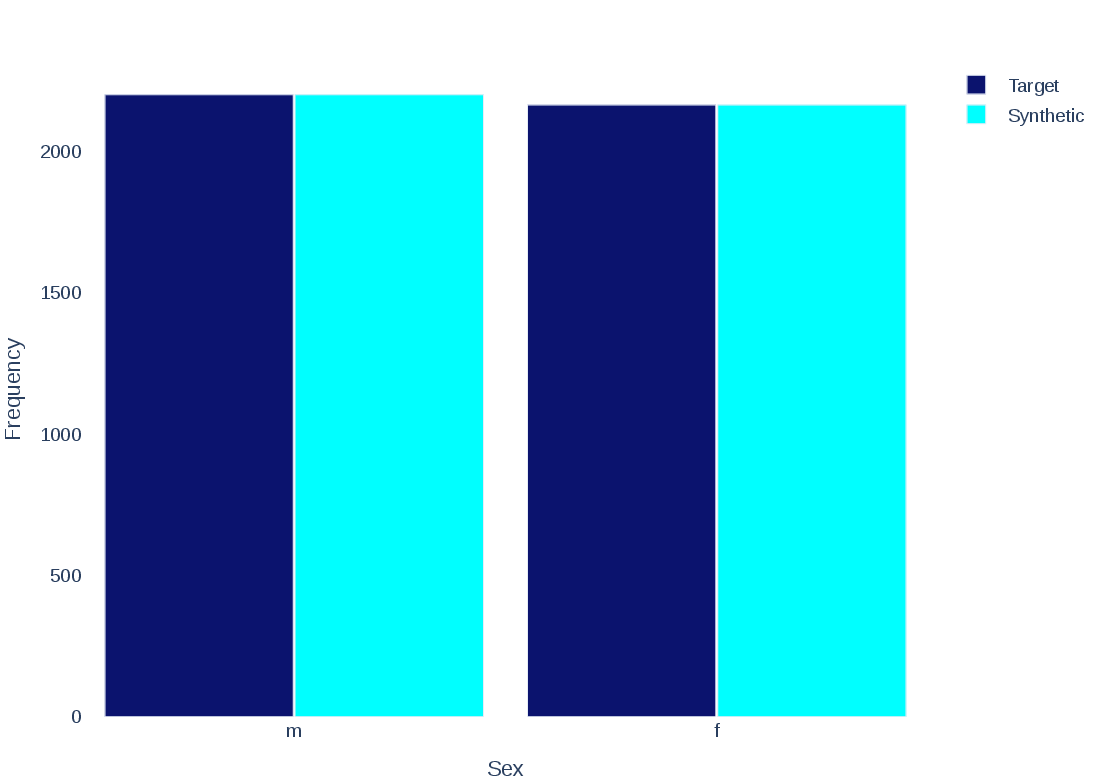}
        \caption{SEX}
    \end{subfigure}
    \hfill
    \begin{subfigure}{0.3\linewidth}
        \includegraphics[width=\linewidth]{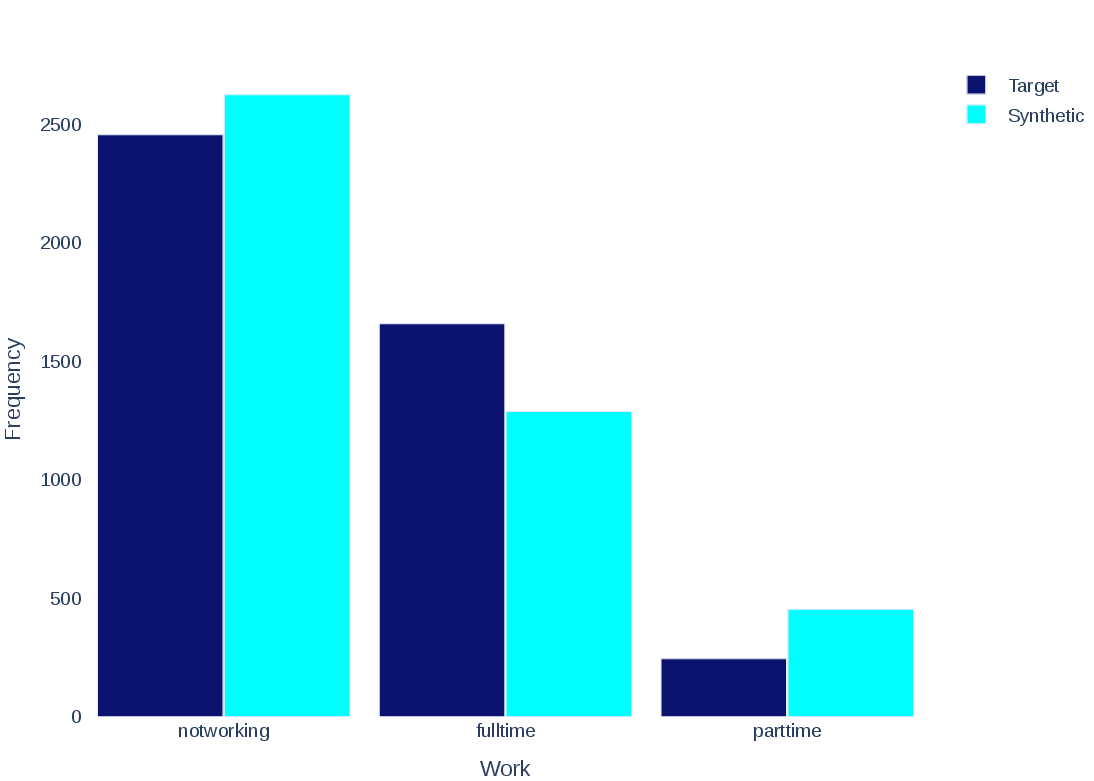}
        \caption{WORK}
    \end{subfigure}

    \vspace{1cm}

    \begin{subfigure}{0.3\linewidth}
        \includegraphics[width=\linewidth]{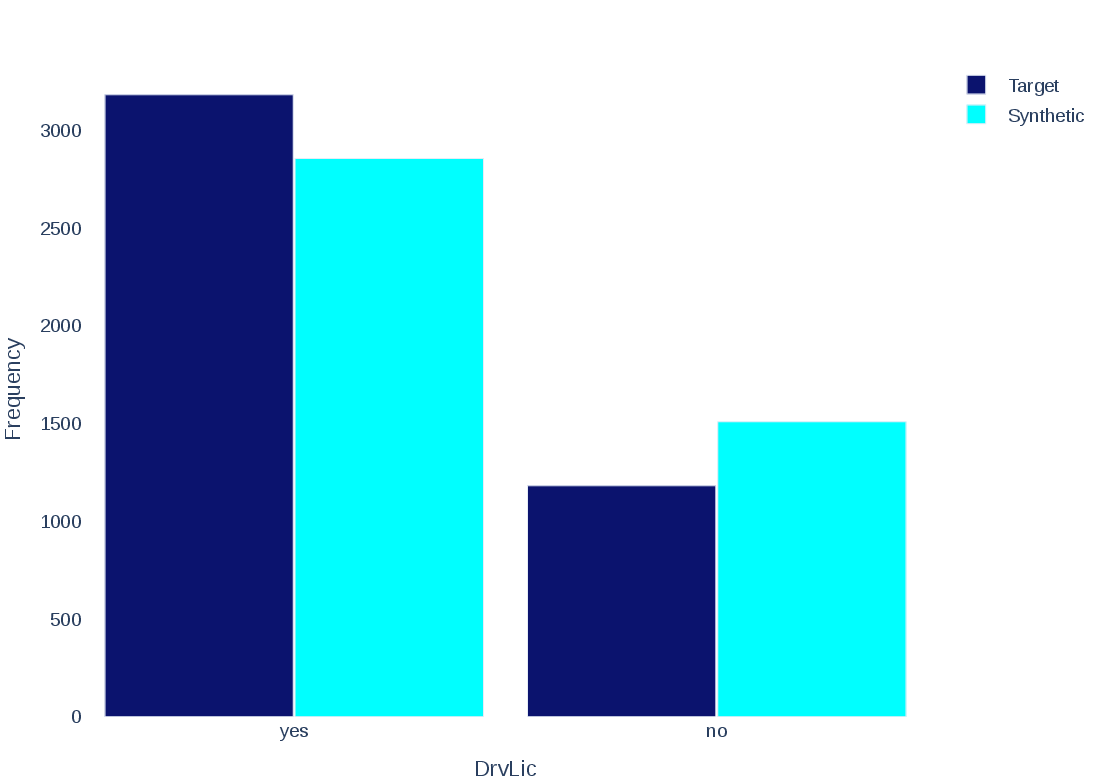}
        \caption{DRVLIC}
    \end{subfigure}
    \hfill
    \begin{subfigure}{0.3\linewidth}
        \includegraphics[width=\linewidth]{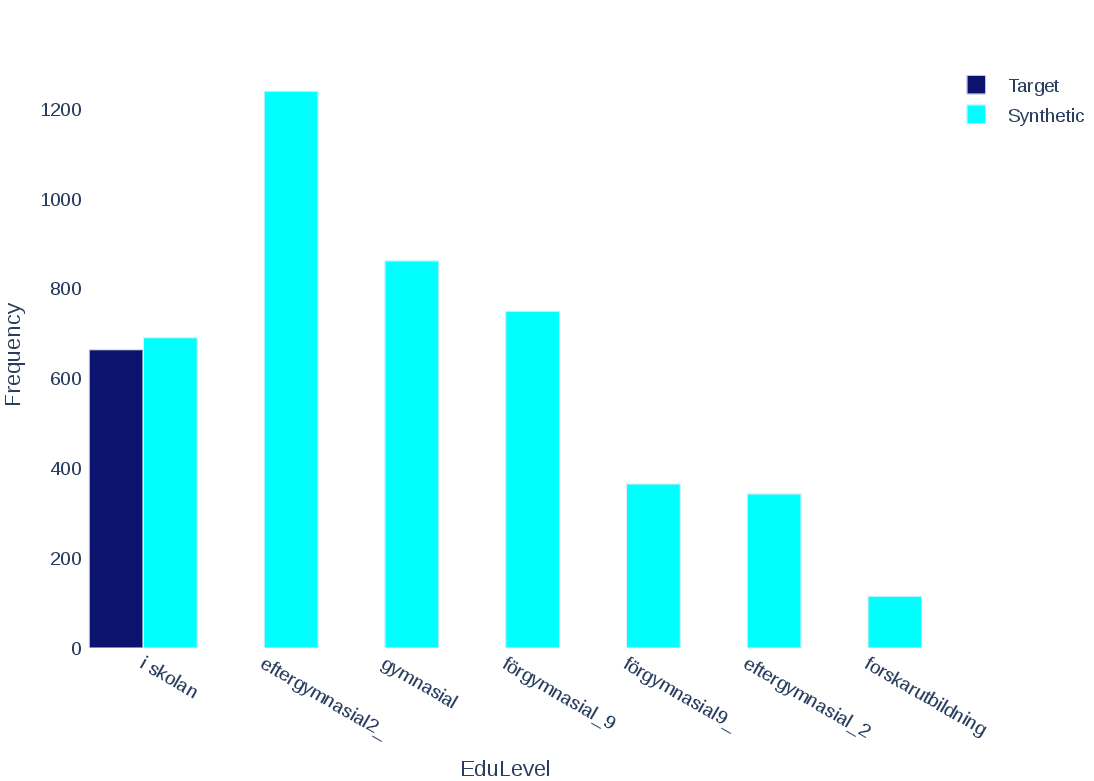}
        \caption{EDULEVEL}
    \end{subfigure}
    \hfill
    \begin{subfigure}{0.3\linewidth}
        \includegraphics[width=\linewidth]{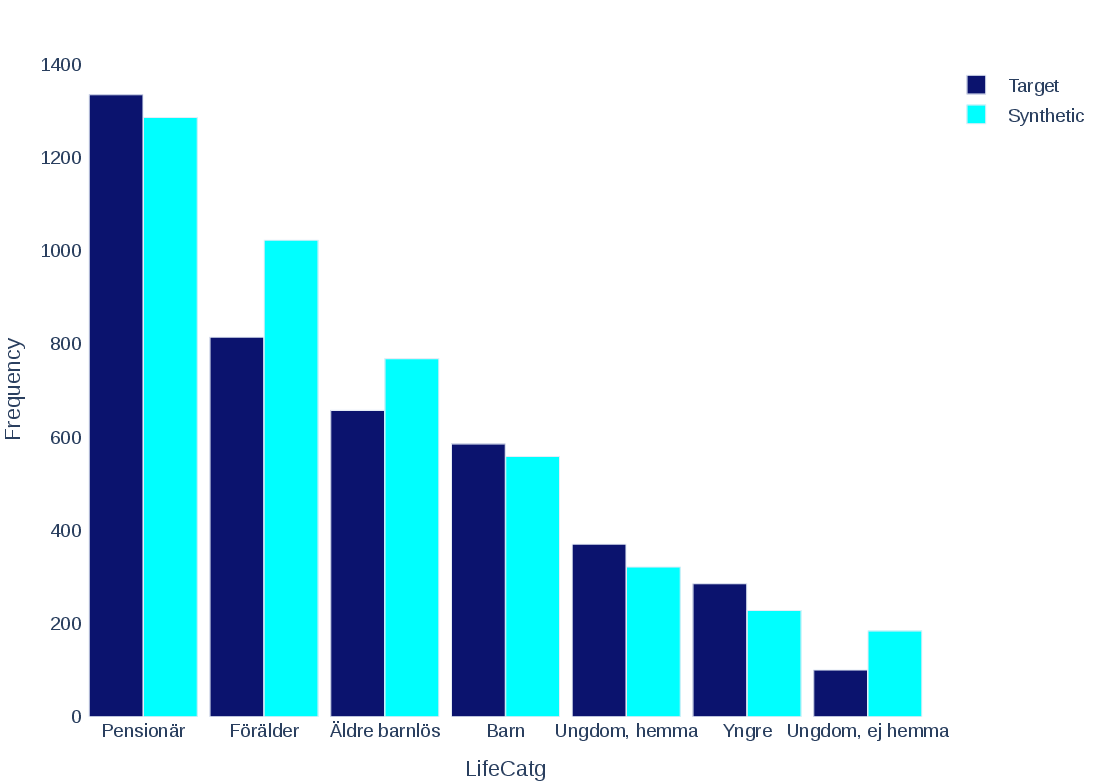}
        \caption{LIFECATG}
    \end{subfigure}

    \caption{Attribute distribution for target and synthetic travel survey data for year 2016.}

\end{figure}

\end{landscape}

\end{document}

%% file: kth_thesis_style_papers.tex
\documentclass[12pt]{article}
\usepackage[T1]{fontenc} 
\fontfamily{bch}\selectfont
\usepackage[skip=8pt plus1pt, indent=0pt]{parskip}
\usepackage{titlesec}
\usepackage{fancyhdr}
\usepackage{setspace}
\usepackage{lmodern}
\usepackage{caption}
\usepackage{subcaption}
\usepackage[none]{hyphenat}

\newlength\titleindent
\newlength\sectionskip
\titleformat{\section}  
  {\vspace{\sectionskip}\fontsize{24}{24}\sffamily\raggedright} 
  {\llap{\parbox[b]{\titleindent}{\thesection\hfill}}} 
  {0em} 
  {} 
  [] 
\titlespacing*{\section}{0pt}{0pt}{24pt plus 1pt}

\titleformat{\subsection}  
  {\fontsize{15}{15}\sffamily\bfseries\raggedright} 
  {\thesubsection} 
  {0.6em} 
  {} 
  [] 
\titlespacing*{\subsection}{0pt}{18pt plus 3pt minus 1pt}{10pt plus 1pt}

\titleformat{\subsubsection}  
  {\fontsize{13}{13}\sffamily\bfseries\raggedright} 
  {\thesubsubsection} 
  {0.6em} 
  {} 
  [] 
\titlespacing*{\subsection}{0pt}{24pt plus 3pt minus 1pt}{12pt plus 1pt}